\pdfoutput=1
\documentclass[11pt]{article}
\usepackage{enumitem}
\usepackage[table]{xcolor}
\usepackage{style/emnlp2022}
\usepackage{todonotes}

\usepackage{times}
\usepackage{latexsym}
\usepackage{graphicx}
\usepackage[T1]{fontenc}
\usepackage{bbold}
\usepackage{array}
\usepackage{mfirstuc}
\usepackage{subcaption}
\usepackage{verbatim}
\usepackage{fancyvrb}

\usepackage{algorithm}
\usepackage{algpseudocode}
\usepackage{amsmath}
\usepackage{graphicx}
\usepackage{soul}
\sethlcolor{green!15}
\usepackage{hyperref}
\hypersetup{
    urlcolor=blue,
}
\usepackage{tabularx}
\usepackage{array}
\usepackage{cellspace}
\newcolumntype{C}[1]{>{\centering\arraybackslash}m{#1}}

\usepackage[utf8]{inputenc}
\usepackage{tabularx}    % Required for tabularx
\usepackage{multirow}
\usepackage{microtype}
\usepackage{inconsolata}
\usepackage{inconsolata}
\usepackage{listings}
\usepackage[most]{tcolorbox} 
\usepackage{pifont}

\usepackage{subfiles} % Best loaded last in the preamble

\usepackage[bb=ams]{mathalfa}
\definecolor{lightgray}{rgb}{0.95,0.95,0.95}

\title{
Large Language Models Struggle to Describe the Haystack without Human Help: A Social Science-Inspired Evaluation of Topic Models
}
\author{
    Zongxia Li\textsuperscript{1,4} \quad Lorena Calvo-Bartolomé\textsuperscript{2} \quad Alexander Hoyle\textsuperscript{3} \quad Daniel Stephens\textsuperscript{4}\\
    \textbf{Paiheng Xu}\textsuperscript{1} \quad
    \textbf{Alden Dima}\textsuperscript{4} \quad \textbf{Juan Francisco Fung}\textsuperscript{4} \quad \textbf{Jordan Boyd-Graber}\textsuperscript{1} \\
    University of Maryland\textsuperscript{1}\quad
Universidad Carlos III de Madrid\textsuperscript{2}\quad
ETH Zürich\textsuperscript{3}\quad
NIST\textsuperscript{4}\\
    \texttt{\{zli12321,paiheng,jbg\}@cs.umd.edu} \quad \texttt{lcalvo@pa.uc3m.es} \quad \texttt{hoylea@ai.ethz.ch}
}

\newcommand{\mallet}[0]{\abr{mallet}}
\newcommand{\mm}[0]{\abr{llm}}

\newcommand{\dvae}[0]{\abr{dvae}}
\newcommand{\ctm}[0]{\abr{ctm}}
\newcommand{\bass}[0]{\abr{bass}}

\newcommand{\lloom}[0]{\abr{ll}{\small oo}\abr{m}}
\newcommand{\topicgpt}[0]{\abr{t}{\small opic}\abr{gpt}}
\newcommand{\topicmistral}[0]{\abr{t}{\small opic}\abr{mistral}}
\newcommand{\topicllama}[0]{\abr{t}{\small opic}\abr{llama}}
\definecolor{softgreen}{RGB}{235,255,235}
\definecolor{bertopiccolor}{HTML}{DD2829}  % Red
\definecolor{ldacolor}{HTML}{3E7FB5}       % Blue
\definecolor{ctmcolor}{HTML}{5AAD50}       % Green
\definecolor{dvaecolor}{HTML}{F98128}      % Orange

\newtcolorbox[list inside=prompt,auto counter,number within=section]{prompt}[1][]{
    colbacktitle=black!60,
    fonttitle=\small,
    coltitle=white,
    fontupper=\footnotesize,
    boxsep=3pt,
    left=0pt,
    right=0pt,
    top=0pt,
    bottom=0pt,
    boxrule=1pt,
    enhanced jigsaw,  % Allows the box to break across pages
    breakable,    
    #1,
}

\newif\ifcomment\commenttrue
\usepackage[a-1b]{pdfx}

\usepackage{framed}
\usepackage{mdwlist}
\usepackage{siunitx}
\usepackage{latexsym}
\usepackage{colortbl}
\usepackage{xcolor}
\usepackage{nicefrac}
\usepackage{booktabs}
\usepackage{fnpct}
\usepackage{amsfonts}
\usepackage[T1]{fontenc}
\usepackage{bold-extra}
\usepackage{amsmath}
\usepackage{amssymb}
\usepackage{bm}
\usepackage{graphicx}
\usepackage{mathtools}
\usepackage{microtype}
\usepackage{multirow}
\usepackage{multicol}
\usepackage{xpatch}
\usepackage{latexsym,comment}
\usepackage[normalem]{ulem}

\newcommand*{\missingreference}{{\Huge \colorbox{red}{?reference?}}}
\newcommand*{\missingcitation}{{\Huge \colorbox{red}{?citation?}}}

\makeatletter
\xpatchcmd{\@setref}{\bfseries}{\missingreference}{}{}
\def\@citex[#1]#2{\leavevmode
    \let\@citea\@empty
    \@cite{\@for\@citeb:=#2\do
        {\@citea\def\@citea{,\penalty\@m\ }%
            \edef\@citeb{\expandafter\@firstofone\@citeb\@empty}%
            \if@filesw\immediate\write\@auxout{\string\citation{\@citeb}}\fi
            \@ifundefined{b@\@citeb}{\hbox{\reset@font\missingcitation}%
                \G@refundefinedtrue
                \@latex@warning
                {Citation `\@citeb' on page \thepage \space undefined}}%
            {\@cite@ofmt{\csname b@\@citeb\endcsname}}}}{#1}}
\makeatother

\newcommand{\gem}[1]{\mbox{\textsc{gem}}}
\newcommand{\abr}[1]{\textsc{#1}}

\newcommand{\lda}{\abr{lda}}

\newcommand{\bertopic}{\abr{bert}{\small opic}}

\newcommand{\hidetext}[1]{}
\newcommand{\ignore}[1]{}

\ifcomment
    \newcommand{\pinaforecomment}[3]{\colorbox{#1}{\parbox{.8\linewidth}{#2: #3}}}

    \newcommand{\prtodo}[1]{\pinaforecomment{lightblue}{pr}{#1}}
    \newcommand{\prtodoi}[1]{\pinaforecomment{lightblue}{pr}{#1}}
\else
    \newcommand{\pinaforecomment}[3]{}
    \newcommand{\prtodo}[1]{}
    \newcommand{\prtodoi}[1]{}
\fi

\newcommand{\smallurl}[1]{ \begin{tiny}\url{#1}\end{tiny}}

\definecolor{lightblue}{HTML}{3cc7ea}
\definecolor{CUgold}{HTML}{CFB87C}
\definecolor{grey}{rgb}{0.95,0.95,0.95}
\definecolor{ceil}{rgb}{0.57, 0.63, 0.81}
\definecolor{UMDred}{HTML}{ed1c24}
\definecolor{UMDyellow}{HTML}{ffc20e}

\begin{document}
\maketitle
\begin{abstract}

A common use of \abr{nlp} by social scientists is to understand large document collections.
Recent data exploration and content analysis have shifted from probabilistic topic models to Large Language Models (\mm{}s).
%A common use of \abr{nlp} is to facilitate the understanding of large
%document collections, 
%with a shift from using traditional topic models to Large Language Models.
% with models based on Large Language Models (\mm{}s) %based models
% replacing traditional machine learning topic models.
%The last decade has seen a shift from topic models for data exploration and content analysis to Large Language Models (\mm{}s) to generate or refine topics. 
%
Yet their effectiveness in helping users understand content in real-world applications remains under explored.
%Yet the effectiveness of using \mm{} for large corpus understanding in real-world applications remains underexplored.
%Yet the ability of \mm{}-based models to help users understand content in real-world applications remains under explored.
%, motivating our empirical evaluation of prompt-based and probabilistic topic models.
%
% We propose \bass{}, an \mm{}-based topic modeling approach that puts users at the center of topic discovery. 
%We conduct an empirical evaluation of prompt-based topic models and probabilistic topic models.
%
This study compares the knowledge users gain from unsupervised \mm{}s, supervised \mm{}s, and traditional topic models across two datasets.
% with topic
% models---including traditional, unsupervised and supervised \mm{}-based
% approaches---on two datasets.
%In this study, we compare \emph{classical} and \mm{}-based topic models by evaluating the knowledge user acquire with the help of topic models on two datasets.
% measuring the knowledge users learn from topic models on two datasets.
%
% \jbgcomment{Be more concrete about what "results" mean.}
%Our evaluation combines standard clustering metrics, response consistency and quality through automatic evaluation, and knowledge acquisition preference through pairwise user preference.
% assessment through pretest/posttest question-answering tests, using both automatic and human evaluations. 
%
%We use an end-to-end framework to measure answer consistency between users and answer quality against gold answers.
%
While unsupervised \mm{}s generate more human-readable topics, their topics are overly generic for domain-specific datasets and do not help users learn much about the documents.
Adding human supervision to \mm{} generation improves data
exploration by mitigating hallucination and over-genericity but requires greater human effort.
Traditional topic models, such as Latent Dirichlet Allocation
(\lda{}), remain effective for exploration but are less user-friendly.
%\mm{}-based methods produce more human-readable topics, and they are more effective for data exploration, but they appear to generate overly generic topics for three domain-specific datasets we tested~\ref{qualitative}.
% but hard to scale up and generate overly generic topics for domain-specific data.
%
%In contrast, traditional topic models like Latent Dirichlet Allocation (\lda{}) can still effectively support data exploration efforts, but is less user-friendly compared to \mm{}-involved topic models.
% remain effective for large-scale data exploration but  
% not as effective as \mm{}-based topic models for data exploration, especially for discovering abstract contents.
%While \mm{}-based approaches produce more human readable topics, they face scalability challenges. 
%
% Traditional topic models like Latent Dirichlet Allocation (LDA) is still effective for large-scale data exploration. 
%
%Furthermore, \mm{}-based models struggle with domain-specific data, often generating overly generic topics less useful for data exploration. 
%
\mm{}s struggle to describe the haystack of large corpora without human help, particularly domain-specific data, and face scaling and hallucination limitations due to context length constraints.\footnote{Datasets are available at \url{https://huggingface.co/datasets/zli12321/Bills}}

\end{abstract}

% \jbgcomment{Meta: this file should be called 2024\_acl\_bass or something}

% \jbgcomment{I don't think the title is as effective as it can be.  Perhaps something more like:

% LLMs cannot yet describe the Haystack with Human Help: A Social Science-inspired evaluation of topic models

% }

% \zongxiacomment{Run Style Checker}

\section{Tools for Corpus Understanding} \label{sec:sections/10-intro} % Exploring and analyzing data strategically and efficiently have become more difficult with increasing data sizes nowadays.
% %
% Probabilistic topic models (\lda{}) use word distributions and neural topic models (\bertopic{}) use pre-trained word embedding to automatically discover latent themes within a corpus~\cite{blei2003lda, bertopicMark}. 
% %
% They have become popular tools for inducing concepts from large unstructured texts by analyzing low-level keyword features from the corpus such as \textit{`health'}, \textit{`insurance'}\dots
% %
% However, those classical topic models produce topics and clusters based on word low-level word signals and often generate topics that are less interpretable and less useful to users~\cite{hoyle2021automated}.
% %

% \jbgcomment{First two \S~s would be better with running example that ties to one of our two datasets.  E.g., User X wants to Y.  Put that in first paragraph then use that in \S~ 2 extensively, the sorts of output you'd get for that problem for TMs / LLMs}

% 
When a researcher approaches a text corpus, they do so with particular
research goals or questions in mind~\cite{krippendorff2004content}:
``Is immigration news coverage focused narrowly on the economic costs
of immigration?''~\cite{annesley2013investigating}; ``How well are the
priorities of the American public reflected in the policy activities
of government?''~\cite{governing}.
To answer these questions, analysts often use corpus analysis
techniques like topic models~\cite[\S~\ref{background}]{boyd2017applications}.
Roughly, these tools structure a document collection by organizing it into
interpretable, high-level categories or topics: newspaper
articles may produce topics relating to the national economy, local
gossip, or sports.
%
%To simulate and measure the learning process facilitated by topic models, we conduct a user study with two datasets: congressional bills and science fiction summaries generated by an \mm{}.
%
For example, consider a federal legislator preparing for a congressional hearing on whether new infrastructure
projects harm ecosystems.
To make informed decisions, they must answer 
%fundamental 
questions like: 
%such as:
%Anyone trying to tackle this issue should have a good answer to the question: 
``What are common policy actions taken by US governments 
%in the US 
to manage land use?'' (\S~\ref{study_setup}).
%effectively
%

% Traditional models and \mm{}s can automatically generate
% topic landscapes without human supervision to facilitate the content exploration process.
% %Traditional topic models and \mm{}-based models like \topicgpt{}~\cite{pham2024topicgpt} automatically generate topic landscapes without human supervision. 
% %
% However, whether these generated topics truly help researchers answer their
% essential questions remains underexplored.
% %
% Despite high automatic evaluation scores, traditional topic models often
% produce outputs that are technically coherent but practically
% unreadable and unhelpful for understanding
% datasets.
% % ~\cite{hoyle2021automated}.
% % \jbgcomment{would be better to avoid self-cites for submission}
% %
% Users have specific information needs that computational models
% struggle to meet, particularly due to hallucination and
% information loss in long or multimodal documents~\cite{liu2023lostmiddlelanguagemodels, vlmSurvey}.
% %
% Critically, most \mm{}-based topic models have been evaluated solely
% through automatic metrics like coverage rate~\cite{lam2024concept} and
% adjusted rand index~\cite{ARI}, without meaningful human validation.
% %
% These metrics fail to capture how social scientists and experts use
% %and derive value from 
% topic models in real-world applications.

%We want to evaluate users' ability to answer similar questions with the help of \abr{ai}.
%
%In contrast, 
While a long line of work has assessed the usability and
interpretability of the topics produced by topic models~\cite{Newman2010AutomaticEO, doogan-buntine-2021-topic},
% \ahintext{could stand to add more probably}, 
comparatively little attention has been given to their ability to
foster human understanding---that is, their capacity to help answer
research questions.
To address this gap, we systematically evaluate both traditional and \mm{}-based topic models, asking: what do humans learn from these models?
Through this evaluation, we compare the strengths and weaknesses of \mm{}s and traditional topic models for exploring large corpora, using a human-in-the-loop study to assess how effectively these models help users understand content (\S~\ref{study_setup}).
%we test comparative advantages and disadvantages of \mm{}s with traditional topic models for large corpora exploration.
%yet
%This work addresses this shortcoming with a more systematic human evaluation of topic models (traditional and \mm{}-based topic models in \S~\ref{background}), asking: what do humans learn from topic models?
%
% \jbgcomment{Add a forward pointer, I don't think we need dataset details in the intro}
%  using two datasets
%We run a human-in-the-loop evaluation to evaluate how effectively traditional topic models and \mm{}s help users understand content using two datasets~\ref{study_setup}.

% user study evaluating how effectively
% topic models help users understand content using two datasets~\ref{study_setup}.
% :congressional bills~\cite{AdlerWilkersonBillsProject} and a synthetic
% dataset of science fiction summaries.
%science fiction %summaries
%generated by an \mm{}.

% \jbgcomment{Forward point to literature review \S~.}

% \jbgcomment{BASS shouldn't be embedded in a paragraph.  We need to have a good introduction of why we use BASS:

% TopicGPT does it all!  why do you need BASS:

% - computer is imperfect
% - user has specific needs
% - IKEA effect / learning}

% \ahintext{Add a paragraph summary of the user study here, framed as a contribution...}

% \jbgcomment{Add a topic sentence to the below paragraph introducing BASS by name}
To personalize and validate topic models so they better adapt to users' specific needs, we also introduce \bass{} (Bot-Assisted Semantic Search), an \mm{}-assisted interactive topic model.
%
% Thus, human validation of topic outputs are more important than automatic scores.
%
\bass{} suggests potential topics while allowing users to iteratively refine them during the topic generation process. By incorporating active learning, it efficiently infers topics for the remaining documents, guiding data exploration while minimizing the need for manual labeling.
We evaluate \bass{} against traditional models and \mm{}s through a user study involving 120 participants across two datasets (\S~\ref{subsec:study_conditions}), follows a two-stage structure: pretest and posttest. Users answer the same set of question using only their prior knowledge in the pretest and with assistance from topics generated by their assigned model in the posttest. 
We include standard cluster metrics, transformer-based pairwise similarity metrics, and manual annotations to evaluate our results.

% The evaluation comprises:
% (1) standard cluster metrics to compare
% generated topic clusters against gold topic clusters, (2) transformer-based pairwise
% similarity to assess response consistency and answer quality, and (3) manual pairwise preference
% annotations to evaluate user preferences between answers generated by
% humans when assisted by different topic models.

%
% We apply these measures to users' responses both before and after they explore the dataset: all approaches improve response quality on dataset-relevant questions compared to baseline users who only have access to a basic document search bar, without additional analytical tools.
%
Traditional models (\abr{lda}) lag
behind \mm{}-based methods for data exploration,
but \mm{}-based methods still have limitations: they are difficult to scale, and no single prompt works universally, making them hard to generalize across diverse datasets.
%they are hard to scale up, and there is no one prompt fits all method
%for \mm{}s, making \mm{}-based methods hard to generalize to diverse datasets. 
%
\S~\ref{qualitative} summarizes the advantages, limitations, and best practices for choosing between traditional topic models and \mm{}s.

 \label{introduction}

\section{Leveraging \mm{}s for Data Exploration} \label{sec:sections/20-background} % \jbgcomment{Rename this section something like "background: tools for corpora understanding".  THis section also needs to answer the question "why not just use QA directly"}

Data exploration is not just about finding \emph{an} answer in a
corpus.
While information retrieval and question answering can help users
find a relevant passage or two in a dataset to answer a specific fact-based question (``What act established civilian government in Puerto Rico?''),
data exploration is not about finding a needle in a haystack---it is
about \emph{describing} the shape and contours of that
haystack~\cite{karpukhin2020densepassageretrievalopendomain,
yang2018hotpotqadatasetdiverseexplainable}.
% \jbgcomment{cite DPR and multihop QA}
%
Instead, these processes follow a systematic approach, like grounded
theory~\cite{chun_tie_grounded_2019}, which involves identifying
themes, connecting information across multiple documents, and applying
complex human reasoning to uncover meaningful insights and reliable
findings.
%Instead, these processes comprise a systematic pipeline like grounded theory~\cite{chun_tie_grounded_2019}, which involves identifying themes within the data, connecting information across multiple documents, and complex human reasoning about interconnected themes for researchers to reach reliable and insightful downstream findings. 
%
For example, to answer the question ``What do policies about land use and
wildlife have in common?'' in the context of congressional policies,
researchers must first identify key topics related to land wildlife management.
%themes and topics about land management and wildlife management first.
%
They then examine relevant documents within those themes, reasoning
through similarities and commonalities to develop a well-supported answer.
%Then they have to read about the relevant documents for those themes, and reason the similarity and common grounds between documents for those themes to reach a reasonable answer. 
%
Unlike question answering, data exploration is iterative, requiring deeper reasoning to discover connection and meaningful insights within the data.
%Data exploration is iterative and reasoning for researchers to uncover meaningful insights and connections within the data, which differs from question answering.

Topic modeling is a widely used tool for assisting researchers with information
retrieval and data exploration, helping to uncover latent topics in document collections~\cite{ABDELRAZEK2023102131}.
%
% With the rising popularity and effectiveness of \mm{}s, using them potential for answering research questions or extracting information seems promising. However, data exploration and content analysis typically require a systematic approach~\cite{chun_tie_grounded_2019}. This process involves comprehensive exploration of datasets, starting with theme identification and often necessitates synthesizing information across multiple documents while reasoning about interconnected themes. Relying solely on \mm{}s throughout this process presents challenges due to their context window limitations, faithfulness issues with large corpora, and difficulties in integrating diverse topics for complex dataset analysis.
%
Since the first probabilistic topic model---probabilistic
Latent Semantic Analysis~\cite[p\abr{lsa}]{hoffmann-99}---many
variants have emerged to support researches in information seeking, data exploration, and forming research questions in education~\cite{SunYan2023}, mental
health~\cite{gao2023discoveringmentalhealthresearch}, social media,
public opinion~\cite{LaureateBuntineLinger2023}, inter alia.
%grounded theory data exploration Different topic models retrieve and
%present information differently, but their ultimate purpose is to
%help researchers understand and learn from a
%dataset.  \jbgcomment{I'm not sure this level of detail is ncessary.
%I think that we could probably abstract this a little bit.  }
%Regardless of how the methods find them, 
Unlike traditional
question answering, topic modeling extracts topics from the documents---identifying
words that co-occur in thematically coherent contexts and
provide an initial topic landscape of the dataset.
While \mm{}s can handle diverse tasks, their data exploration
capabilities still require an iterative, systematic pipeline rather than simple
prompting.
Recent topic modeling approaches use \mm{}s to generate more human readable and
descriptive topics~\cite{openai2024gpt4,touvron2023llama}.
%
% TopicGPT~\cite{pham2024topicgpt} samples documents to induce topics, merge topics, and then traverse documents to assign topics and cluster documents.
%
% \jbgcomment{I'm not sure models we don't compare against should be in this section; perhaps move to the end.  But if they do stay here, cite BERTopic}
% \citet{reuter2024gptopic} introduce a method akin to BERTopic but allowing for a fixed number of topics and using agglomerative clustering, defining topics through titles, descriptions, assigned documents, and document embeddings, with RAG enabling dynamic interactions via a chat interface. 
%
% LLooM~\cite{lam2024concept} uses \mm{}s to extract key sentences from individual documents, summarize documents, then use clustering algorithms to cluster summarizations and prompt \mm{}s to generate topics for all clusters.
%
% The above mentioned classical topic models and prompt-based topic models are fully automated with no human intervention to participate in the process of topic generation.
\topicgpt{}~\cite{pham2024topicgpt} and \lloom{}~\cite{lam2024concept} represent topics with more intuitive short descriptions such as \textit{Land Management: Involves policies and actions related to the use, regulation, and conservation of land and natural resources}. 

Thus, we pose a new question: are \mm{}s ready to replace traditional topic models for describing the haystack of a corpus?
%
% \jbgcomment{I think we need to make it clear that the replacement is in the context of these corpus-exploration analyses}
%
To address the motivation, we design an end-to-end evaluation study to
compare \abr{lda}, \mm{}-based methods, and \bass{} to study whether
human supervision can make \mm{}s a more
effective tool for large corpora understanding.
%
% We study whether human involvement is still necessary to supervise \mm{}-based topic models and whether human involvemen
% \bass{} allows users to review only a small set of documents and then revise the automatically generated topics to form a topic set for the entire dataset.
% \zongxiacomment{Add a little bit BASS description here.}
%
% To evaluate the usefulness of these topic models for data exploration, we study the following research questions: (1) Can we replace traditional topic models with \mm{}s for data exploration? (2) Does human involvement in the topic generation process have a positive or negative impact on user experience?

% We use an \mm{} in interactive topic modeling to alleviate this burden by leveraging \mm{}s to provide topic suggestions, we enable users to generate topics under human supervision, reducing the need for complete manual topic creation while maintaining control over the process.
%
% Fortunately, with existing interactive topic model techniques and prompt-based topic models, we can leverage interactive topic models such that humans collaborate with an \mm{} and active learning classifier to quickly learn knowledge about a dataset.
 \label{background}

% \jbgcomment{I'm not sure that BASS is introduced prominently enough.  I think that should probably happen here, perhaps in its own section.  Otherwise, it's going to get lost as a contribution.}

\section{Setup and Evaluation} \label{sec:sections/30-study-setup} %In this section, we introduce the basic setup of the user study, including the datasets we use, control groups, and the pre-test and post-test questions.

% \jbgcomment{The section should be renamed something like "Measuring what Users Learn from Interacting with Data"}

% \jbgcomment{I'm not 100\% sure, but I \emph{think} this should come before the previous section}

% \jbgcomment{Can we rename / renumber the files to match the paper sections}

% \jbgcomment{Let's bring back the running example here to motivate the introduction: our hypothetical user wants to understand land management, how can we measure how well they accomplished that goal}

Suppose a researcher wants to
understand US policy on land management. How can we measure their understandings of the question from the corpus?
%Suppose our hypothetical legislator from the introduction wants to understand common policy actions taken by governments in the United States to manage land use for a legislation draft, how can we measure how well they accomplish the goal effectively?
%
% Our goal is to conduct an end-to-end evaluation of content analysis by assessing the effectiveness of various topic modeling tools.
%Our goal is to do an end to end evaluation of content analysis-- how effective is various topic model tools? 
%
We need to ensure answers are faithful, comprehensive, and link
multiple documents while maintaining consistency.
%We want to ensure the answer to be faithful and comprehensive that links multiple policies, while insights are consistent independent of specific legislators.
%
With this in mind, we use a pretest--posttest method: users
answer questions before and after interacting with the models.
The better they answer the questions, the more the models helped them
ingest the key themes of the dataset.
%, asking the same questions before and after users interact with the topic model and datasets. 
%
The pretest is critical to control for users with prior expertise (or
skill in making up convincing-sounding answers) who can answer
question well without assistance.

\subsection{Datasets}\label{sec:datasets}
We use two datasets for our evaluation, \texttt{Bills} and \texttt{Sci-Fi}, chosen because they are less likely to be part of \mm{}'s pretraining data. 
%We evaluate on two datasets, \texttt{Bills} and \texttt{Sci-Fi}. We choose them because they are less likely to be from \mm{}'s pretained data.

% After running synthetic experiments, we chose these to ensure they were absent from the \mm{}'s pretraining data and allowed measurable user knowledge gain (see \S~\ref{appen:synthetic_experiments} for details).

\paragraph{\texttt{Bills}}
is a standard
benchmark for topic
models~\cite{AdlerWilkersonBillsProject}.
% both traditional and prompt-based
%
The dataset contains 32,661 bill summaries from the
110\textsuperscript{th}--114\textsuperscript{th} U.S. Congresses,
categorized into 21 top-level and 112 secondary-level topics, and we collect 11,327.
%(\textit{health, education, defense, trade, etc.})  and 112
%secondary-level topics (\textit{public lands, electricity, etc.}),
%from which we collect \num{11,327}.

%To verify the generality and robustness of our conclusion, we use the Bills dataset~\cite{AdlerWilkersonBillsProject}, which has been used by traditional topic models like \lda{}~\cite{blei2003lda}, and prompt-based models such as \topicgpt{}~\cite{pham2024topicgpt}, and \lloom{}~\cite{lam2024concept}, and can provide us an easy and straightforward comparison with existing topic models.

%, which has become a standard benchmark for topic model.% that contains hierarchical labels refined through the years with inter-annotator agreement.
%
%The dataset contains 21 generic top level topics (\textit{health, education, defense, trade, etc}) and 112 secondary-level topics (\textit{public lands, electricity, etc}). 

% 
%We collected 11,327 documents from the Bills data.

\paragraph{\texttt{Sci-Fi}}%Synthetic 
Inspired by \citet{lam2024concept}, we use \textsc{LLaMA-3 70B} to
generate a synthetic dataset of two thousand imaginary science fiction story summaries.
Our goal is to create a controlled
dataset with predefined questions, answers, and themes that probe topics requiring cross-document reasoning--insights difficult to
extract from individual documents and rarely available in real-world datasets. For generation, we select sixteen ground truth Sci-Fi themes that require minimal expertise, then prompt the \mm{} with these
themes and a set of question-answer pairs aligned with our research
objectives (see \S~\ref{appendix:synthetic} for more details on the generation process).

\subsection{Question Generation}
Since our goal is not 
information retrieval or question answering,
our questions must require users to synthesize information from multiple
documents.
Ground truth answers are also necessary for evaluating knowledge
gain. 
In the \texttt{Sci-Fi} dataset, this is ensured by a controlled generation process—documents are created from predefined question-answer pairs that require users to gather information and infer themes across multiple documents.
Am example question is \textit{How does the presence of an unknowable alien intelligence affect the psychological state of the crew aboard space stations?}, which falls under the theme \textit{communication and understanding: investigating the challenges between human and non-human intelligence interaction.}
%
%\jbgcomment{Review em and en dashes, en dash was used incorrectly}
%
For \texttt{Bills}, authors review gold-standard topics and a substantial set of associated topics before drafting and collaboratively refining questions. 
These questions are designed to require reading and analyzing multiple documents within the same or similar topics.
%along with a substantial number of documents associated with selected topics, then draft and collaboratively revise questions whose answers require the reading and analysis of multiple documents under the same or similar topics.
%
All questions and reference answers are finalized before user studies to ensure clarity and fairness (see full questions in \S~\ref{tab:questions}).

\subsection{\textit{Metrics} and \texttt{Evaluation}}

We use two automatic metrics and one human evaluation to assess the
quality of user responses generated with the aid of topic models.

\paragraph{\textit{Answer consistency.}}
%
%\jbgcomment{If the paragraph is not integrated into the next sentence put a period at the end.}
%
We assess answer consistency by computing the pairwise cosine similarity between transformer embeddings for each pair of user answers to the same question. Higher similarity indicates that users are able to use the tool effectively to obtain consistent results.~\cite{tang2022investigatingcrowdsourcingprotocolsevaluating}.
%\paragraph{Pairwise similarity} quantifies the consistency and similarity of users' answers. 
For a question answered by \(K\) users, the consistency score is
\begin{equation}
S = \frac{2}{K(K-1)} \sum_{i=1}^{K-1} \sum_{j=i+1}^K \cos(a_i, a_j)
\end{equation}
where $\cos(a_i, a_j)$ is the cosine similarity between the embedding representation of the answers provided by two users, and $\frac{2}{K(K-1)}$ normalizes the sum to account
for the number of answer pairs.

\paragraph{\textit{Answer quality.}}
We evaluate the quality of user responses by comparing them to the
reference (gold) answers. The similarity is calculated as the cosine
similarity between the transformer embeddings of the user response and
the gold answer, a common way to evaluate how closely users' responses align with the ideal response~\cite{li-etal-2024-pedants, zhang2020bertscoreevaluatingtextgeneration}

\paragraph{\texttt{Pairwise response preference.}}
%
%Even though 
While gold answers approximate the expected response, they are not the only acceptable ones. %answers. 
%
% \jbgcomment{perhaps we should have different typography for the evals
  % / metrics} 
To complement the {\it Answer quality} metric, we hire annotators to assess response quality using a fixed
rubric.\footnote{The scoring rubric was initially developed by two
social science experts and refined iteratively. It evaluates how well
answers synthesize information from multiple documents to address the
question and penalizes hallucinated content or references outside the
corpus (\S~\ref{appendix:rubric}).}
Assigning precise scores to responses is challenging, so evaluators
are asked to compare pairs of responses and select the better
one.
We then apply the \citet{Bradley1952RankAO} model to compute pairwise preference strength, then rank models based on evaluator-perceived answer quality (Examples in \S~\ref{sec:domain_specific_data}).
 \label{study_setup}

\section{Methods} \label{sec:sections/40-method} % \jbgcomment{Begin this section by saying something like 

% "This section outlines the N techniques which we compare in our user study in Section X"
% }
This section outlines the four methods compared in our study.

\subsection{Study Conditions}\label{subsec:study_conditions}
%We want to measure the information gain and knowledge learned by using various topic model tools and comparing them on content analysis.
% \jbgcomment{This section is talking about things that are not control groups.  Perhaps have a section called "conditions"?

% Or perhaps "control group" is used incorrectly: 
% \url{https://en.wikipedia.org/wiki/Treatment_and_control_groups}
% }

%
% To measure the effectiveness of topic models helping the data exploration process,
%\jbgcomment{Put citations next to name}
% We divide user study participants into four groups, where each group uses a different topic model to explore the data---\lda{}~\cite{blei2003lda},\topicgpt{}~\cite{pham2024topicgpt}, \lloom{}~\cite{lam2024concept} (unsupervised), and our new framework, \bass{}.

We assign participants to four groups, each using a different approach to explore the data: one traditional topic model, two recent \mm{}-based models, and our framework. Specifically, we compare \lda{}~\cite{blei2003lda} against \topicgpt{}~\cite{pham2024topicgpt} and \lloom{}~\cite{lam2024concept} (unsupervised), as well as our approach, \bass{}. 
%

\begin{figure*}[!t]
    \centering
    \includegraphics[scale=0.8]{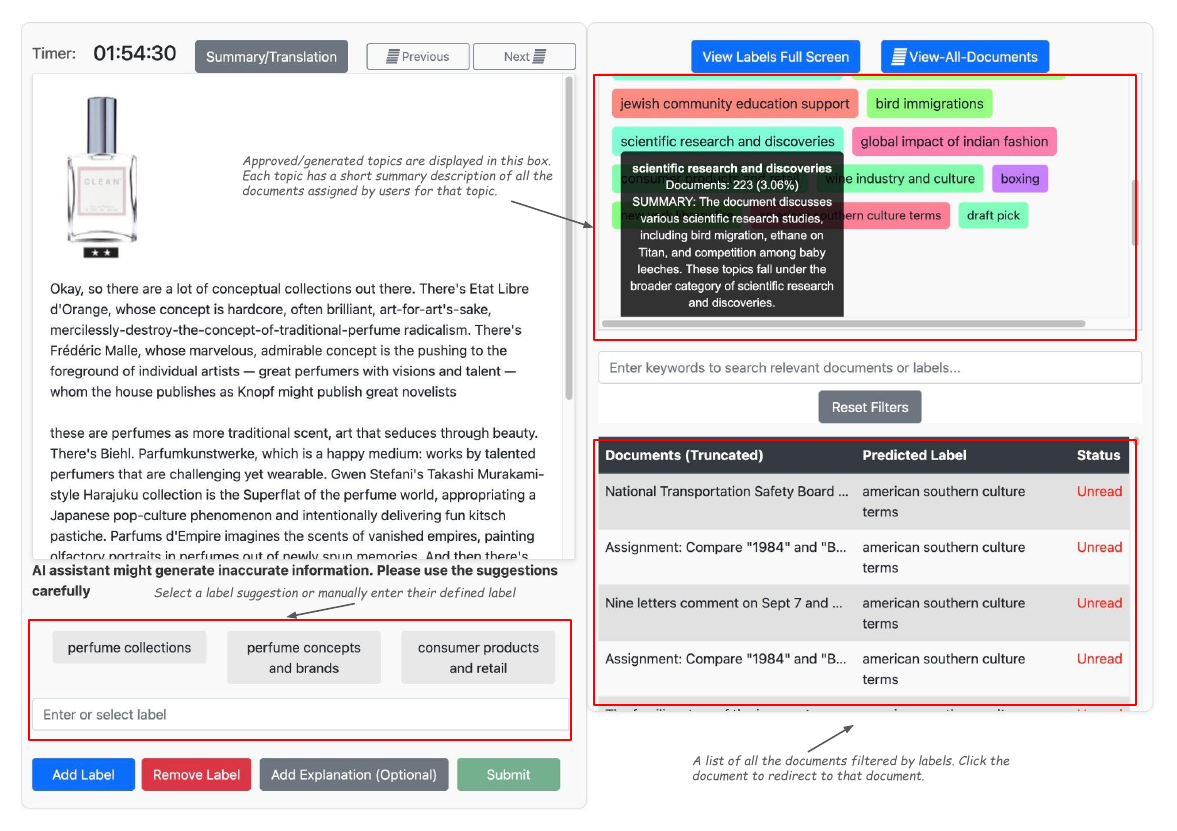}
    \caption{
    % Topic-generation process for \bass{} users. From top to bottom, and left to right: users are presented with a document and its summary (automatically generated by an \mm{}), three \mm{}-suggested labels (from which they can select one or input a new one), the already generated topics (hover to see the number of assigned documents), and a search bar to find similar documents using keyword and TF-IDF search.
    Users are shown selected documents and topics generated by topic models or topics they have added (hover to see the number of assigned documents). In \bass{}, the UI presents three \mm{}-suggested labels (from which users can select one or enter a new one), but they start with empty set of topics. The search bar allows users to find similar documents using keyword and \textsc{tf-idf} search. Users in the baseline group do not receive \mm{}-suggested labels nor topics generated by a topic model, but can create and assign new labels manually.
    }
    \label{fig:bass_topic_generation}
\end{figure*}

%We use \abr{lda}, TopicGPT, LLooM~\cite{blei2003lda, pham2024topicgpt, lam2024concept}, and \bass{}. 
%
While \lda{}, \topicgpt{}, and \lloom{} (unsupervised) generate topics automatically without human intervention, \bass{} starts with no predefined topics.
%initially presents no topics to the user.
%
%Instead, the user generates topics by supervising the \mm{}
%suggestions and reviewing documents
Instead, users generate topics by reviewing representative documents selected via active learning and considering topics suggested by an \mm{} (Fig.~\ref{fig:bass_topic_generation}). 
Users can adopt these suggestions or propose new ones.
Based on user inputs, the active learning classifier clusters documents and infers labels for the remaining ones, reducing annotation costs.
%then uses user inputs to cluster documents and infer labels for the rest to reduce annotation cost.
%

% \jbgcomment{If you're going to abbreviate Figure, create a macro so we can undo it later}
%(Figure \ref{fig:bass_topic_generation}).
%
%Then an active learning classifier trained the users' labels
%infers labels for the remaining documents.
%

%\jbgcomment{These things don't belong together in the same paragraph.  This should just be about BASS.  The questions should be in a different paragraph}
%
The study consists of two phases: pretest and posttest (\S~\ref{study_setup}). 
In both, users answer the same set of questions related to their assigned dataset. 
During the pretest, users rely on their own knowledge, allowing the identification of participants with significant prior knowledge of the questions, and their test-taking ability. 
In the posttest, users are assisted by topics generated by the assigned topic model---except in \bass{}, where users create the topics. 
In all conditions, a search bar with string matching and \textsc{tf-idf}\footnote{\url{https://www.npmjs.com/package/ts-tfidf}} is available during the posttest.
%In the pre-test stage, all users answer the same set of questions from their assigned dataset to evaluate their baseline expertise, helping to identify outlier participants with significant knowledge about the questions.  
%
%In the post-test stage, unsupervised groups use model-generated topics to answer the same questions, while the \bass{} creates their own topics. 
%
%Users need to review representative documents selected via active learning, review or revise \mm{} suggested topics, and rely on an active learning to cluster documents using their inputs.
%
%All users can search for relevant documents using a search bar with string matching and \textsc{tf-idf}.  
%
% After answering the questions, participants complete a survey.  
Next, we discuss the details of the topic models used.

%and do not require prompting.
%\jbgcomment{rather than prompting, I think it would be better to say that they are completely unsupervised after the vocabulary and number of topics are chosen}
\paragraph{Traditional Topic Models} are unsupervised, once the vocabulary and number of topics are set, representing topics as keywords (e.g., \textit{`health'}, \textit{`insurance'}). 
%representing topics as keywords (e.g., \textit{`health'}, \textit{`insurance'}\dots) given the vocabulary and number of topics.
%
Here, we categorize both probabilistic and neural topic models as traditional topic models.

%Probabilistic and neural topic models (\lda{}) are both considered keyword-based traditional topic models in our case.
%
The exemplar of probabilistic topic model, \lda{}, uses a Bag-of-Words representation to induce latent topics, assigning documents with frequently co-occurring words to the same
topic. 
Neural topic models, in contrast, rely on neural architectures~\cite{prodLDA,burkhardt2019decoupling} to offer richer topic representation, aiming for improved scalability and higher topic quality. They can also be easily integrated with pretrained embeddings~\cite{bianchi2020pre,kitty_CTM}.

%use pretrained  embeddings such as Word2Vec~\cite{word2vec}, sentence-transformers representations to assign similar documents to the same topic.
%
%The topic vector representations for probabilistic and neural topic models are the same, 
Both probabilistic and neural topic models use the same topic vector representations. Given a set of documents and a predefined number of topics\footnote{We use \num{65} as the topic number, which is the average number of topics generated by \lloom{} and \topicgpt{}}, each document can be represented as a vector:
\begin{equation}
\theta^d = \{\theta^d_1, \theta^d_2, \ldots, \theta^d_K\},
\end{equation}
where $K$ is the number of topics, and $\theta^d_i$ is the probability of document $d$ assigned to topic $i$.
We denote the collection of all topic distributions across the corpus as \(\Theta = \{ \theta^d \}_{d=1}^{D}\).
% \begin{split}
%     \boldsymbol{\theta}_d  &=  [\theta_{d,k}, k = 1, \dots, K] \\
%     &\text{with} \: \sum_{k=1}^K \theta_{d,k} = 1, \forall d, \: d=1,\dots,D
% \end{split}

% \(\theta_{d,k}\) is the probability of topic \(k\) for document \(d\)

%
We select \lda{} as the representative traditional model because prior research identifies it as the most stable, with recent neural topic models failing to outperform it and thus recommending it for content analysis~\cite{hoyle-21}. Synthetic experiments comparing \mallet{} \lda{}~\cite{mallet}, Contextualized Topic Models~\cite{kitty_CTM}, \bertopic{}~\cite{bertopicMark}, and the Dirichlet Variational Autoencoder Model~\cite{burkhardt2019decoupling} across three datasets further support these findings (\S~\ref{appen:synthetic_experiments}).

\paragraph{Unsupervised \mm{}-based Exploration.} 
% generate topics by iteratively prompting an \mm{}.
%
Unlike probabilistic models, \mm{}-based models build a distributed representation of the document and then generate a label from that
representation.
%Unlike traditional approaches, these models use documents' semantic
%meanings to generate and assign topics.
%\jbgcomment{I'm not sure that I agree with this.  Perhaps something like:
%Unlike probabilistic models, \mm{}-based models build a distributed
%representation of the document and then generate a label from that
%representation.  }
%
We select \topicgpt{}~\cite{pham2024topicgpt} and \lloom{}~\cite{lam2024concept} as representative exemplars of this group.
%\mm{}-based topic models generate topics by prompting an \mm{} iteratively until the topics converge.
%
%Unlike traditional models, prompt-based models generate topics based on semantic information from the documents and cluster documents based on their semantic information.
%

\topicgpt{} prompts an \mm{} with fixed examples from the dataset,
merges and refines similar topics (transformer cosine similarity $(< 0.5)$), and then assigns the refined topics to all documents. \lloom{}
first prompts an \mm{} to extract key sentences, clusters them using
cosine similarity, and then prompts the \mm{} again to summarize and
generate topics.

%For a set of given documents, \topicgpt{} samples $N$ documents to generate topics by prompting an \mm{} with fixed demonstration examples.
%
%Next \topicgpt{} prompts an \mm{} with a similar topics (transformer cosine similarity < 0.5) to merge and refine topics.
%
%The last step is to prompting an \mm{} to assign the refined topics to all the documents in the dataset.
%
%On the other hand, LLooM uses a different appraoch to prompt \mm{} to generate topics by first extracting key sentences from the documents, then use transformer cosine similarity to cluster relevant extracted sentences.
%
%Then LLooM uses an \mm{} to summarize clustered extracted sentences to synthesize and generate topics.
%

%Prompt-based topic models require access to a reasonable strong \mm{} that can extract and generate reasonable topics from the data.
%
Prompt-based topic models provide more descriptive topic descriptions,
like \textit{Trade: focuses on the exchange of goods} for
\topicgpt{}. 
However, they require numerous \mm{} calls (potentially
expensive for larger datasets), and their practical effectiveness for
user applications has not been fully evaluated.

% they are non-probabilistic and lack the document-to-topic and word-to-topic distributions of traditional models. Moreover, they require multiple \mm{} calls.
%Moreover, a key limitation of these models is that they require multiple \mm{} calls.
%Prompt-based topic models generate descriptive topics that summarizes a document cluster-- \textit{Trade: discusses the exchange of good}.
% The topic representation of topics for prompt-based topic models are a short phrase with a description that describes a common theme documents under the phrase share-- \textit{Trade: discusses the exchange of good}.
%

%On advantage of prompt-based method is that each document can have one more more associated topics, but the disadvantage is that we lose the probability of the document belonging to each topic, which is the $\theta$ in classical topic models. 

\paragraph{\bass{}: Bot-Assisted Semantic Search.}
% \jbgcomment{I think this section should both be earlier and should be called something like 

% BASS: Adding Generative Deep Learning to Interactive Topic Models
% }
% While algorithms and models
While algorithms often generate imperfect topics misaligned with user
intents, \citet{dietvorst2016overcoming} show that users are willing to use
imperfect systems if they can control and modify.
Mixed-initiative
interaction~\cite{horvitz1999mixed-initiative} enables human and \abr{ai} to collaborate, using their complementary strengths to
enhance accuracy and productivity.
%Mixed initiative interaction theory posits that humans and AI should collaborate as complementary agents, each contributing their suitable capabilities to enhance overall accuracy and productivity~\cite{horvitz1999mixed-initiative}.
%
Building on these insights, we use an \mm{} agent to generate topics
while allowing users to maintain control and modify them.
Users supervise the generated topics, approving or revising them
through an interactive process.
%Extending the insights, we want an AI agent generate topics that it can easily recognize, but the preserve the agency of the user.
%
% Unlike fully unsupervised topic models like \topicgpt{}, we focus on increasing user understanding of the topic generation process  to improve interpretability.
%Unlike fully unsupervised topic models like TopicGPT, we want to improve user visibility into the topic generation process to improve topic transparency and interpretability.
%
% Our goal is for users to gain new insights into a dataset through the interactive process.

% Thus, we integrate generative language models into interactive topic modeling, maintaining human supervision by allowing users to accept or modify topics generated by \mm{}s to assist topic generation~\cite{li-etal-2024-improving}. 
%

However, having human to go through all the documents with an \mm{} can be time-consuming and labor-intensive.
Thus, we use active learning~\cite{active-learning}, leveraging user-labeled documents to train a classifier that can infer topics and cluster unseen documents to reduce users' workload.
Specifically, we use the \(\Theta\) from a trained \lda{} model to represent document clusters and similarities, combined
with \textsc{tf-idf} encodings, as features for the active learning
classifier~\cite{alto, li-etal-2024-improving}.
During the topic-generation process, users receive a summary and three
candidate labels an \mm{} for each document, which they can approve,
revise, or reject (Fig.~\ref{fig:bass_topic_generation}).
The labeled documents serve as training data for active learning to update and refine the
topic model's document distributions.\footnote{We use incremental
learning to train the classifier. If a new label class is created, the
classifier is reinitialized and retrained. The prompt template to
generate suggested topics is in Appendix
Fig.~\ref{fig:bass_prompt}.}
The classifier then generalizes these {\it ``fine-tuned''} \abr{lda}
topic representation to unseen documents.\footnote{Classifier training details in \S~\ref{app:synthetic_experiment_setup}}

 \label{method}

\section{Results and Analysis} \label{sec:sections/50-results} % \jbgcomment{For all of the plots, we should capitalize the conditions in the legends}

\begin{table*}[t]
    \centering
    \tiny
    \renewcommand{\arraystretch}{1.2}
    \setlength{\tabcolsep}{3.5pt}  % Slightly reduced spacing to fit new column
    % Define row colors
    % \rowcolors{2}{gray!15}{white}
    \begin{tabular}{l cccccc cccccc}
    \toprule
    \multirow{2}{*}{\textbf{Metric}} & 
    \multicolumn{6}{c}{\textbf{\texttt{Bills}}} & 
    \multicolumn{6}{c}{\textbf{\texttt{Sci-fi}}} \\
    \cmidrule(lr){2-7} \cmidrule(lr){8-13}
    & \textbf{\abr{lda}} & \textbf{TopicGPT} & \textbf{TopicMistral} & \textbf{TopicLLaMA} & \textbf{LLooM} & \textbf{\bass{}} &
      \textbf{\abr{lda}} & \textbf{TopicGPT} & \textbf{TopicMistral} & \textbf{TopicLLaMA} & \textbf{LLooM} & \textbf{\bass{}} \\
    \midrule
    Purity & \textbf{0.70} & 0.52 & 0.75 & 0.53 & 0.23 & 0.54 & 
             \textbf{0.63} & 0.35 &  0.44 & 0.26   & 0.26 & 0.28\\
    ARI & 0.27 & 0.23 & 0.09 & 0.21 & 0.16 & \textbf{0.45} & 
          \textbf{0.18} & 0.14 &  0.08 & 0.11 & 0.11 & \textbf{0.18} \\
    NMI & 0.47 & 0.42 & 0.13 & 0.39 & 0.14 & \textbf{0.54} & 
          0.56 & 0.18 & 0.06 & 0.17 &  0.19 & \textbf{0.69} \\
    Num Topics & 65 & 73 & \textbf{317} & 118 & 44 & \(\bar{u}\)=46 & 
          65 & 4 & 33 & 21 &  \textbf{269} & \(\bar{u}\)=53 \\
    \bottomrule
    \end{tabular}
    \caption{
       Traditional, automatic, label-centric clustering metrics (Purity, ARI, NMI) and number of topics for each model on the \texttt{Bills} and \texttt{Sci-Fi} datasets. In \lda{}, the number of topics is predefined, while in fully prompt-based models (\topicgpt{} and \lloom{}) it is discovered automatically. For \bass{}, we report the average number of topics generated across a 15-user sample per dataset with standard error. \bass{}, our proposed method, achieves competitive results, outperforming other models on most metrics. \lda{} performs well overall, \topicgpt{} generates generic topics---especially on the \texttt{Sci-Fi} dataset---and \lloom{} shows lower scores. \topicmistral{} and \topicllama{} uses the same pipeline as \topicgpt{}, but with local Mistral-7B-Instruct and LLaMA-2-70B. Small local \mm{}s have the lowest cluster scores than large local \mm{}s, which is comparable but still worse than close-source GPT models.
    }
    \label{tab:cluster_metrics}
\end{table*}

% \subsection{Participant Recruitment}
We recruit \(15\) users from Prolific for each control group and dataset, totaling \(120\) users. 
The recommended study time was \(45\) to \(60\) minutes, and each user could participate only once.\footnote{All annotators have an approval rate of at least $99\%$ and a minimum of \(30\) previous submissions.}
%
%Each user can only participate the study only once.
%We hire 15 users from Prolific for each group for the Bills and synthetic news dataset, a total of 120 users.
%
%We had different topical requirements for the two datasets: 
Annotators have social science background for \texttt{Bills} and English literature background for \texttt{Sci-Fi}.
% The datasets had different topical requirements: a social science background for \texttt{Bills} and proficiency English for \texttt{Sci-fi}.
%All the annotators participating in the Bills dataset requires to have a social science background or degree and no specific expertise is required for the synthetic news dataset.
%
%We filter annotators to have at least $99\%$ approval rate and 30 previous submissions.
%
% \jbgcomment{I'd move this to the ethics section if we're short on
%   space, but I'd put the \$17 first.  In any event rewrite so it's not
%   passive voice.}
%
% Participants are compensated a base rate of
% $\$6.5$, which could increase to $\$17$ per hour if their answers are
% deemed unlikely to be AI-generated and showing they have done the
% task. We use a fine-tuned \textsc{RoBERTa}~\cite{sivesind_2023} to
% reject users likely submitting AI-generated responses.
% %
% They are provided with task instructions at the beginning
% (\S~\ref{tab:congressional-bills-study},~\ref{tab:news-post-analysis})
% and can quit the study at any time.
% %
% Upon completion, users are prompted with two survey questions to rate
% their experience using the tool (\S~\ref{fig:survey_ratings}).
%
The rest of this section analyzes the results from the user study.

\subsection{Topic Clustering Metrics}
Evaluating cluster similarity is challenging~\cite{NIPS2002_43e4e6a6}.
Rather than comparing text-based topic summaries---since \lda{} produces keywords while \topicgpt{} and \lloom{} generate more fluent but not necessarily more useful phrases---we focus on cluster assignments.
%phrases, which may be more fluent but not necessarily more useful, we focus on cluster assignments.
%
Specifically, we evaluate document assignments to each topic (cluster) by comparing them to a 
%based on a 
ground truth partition, measuring the similarity between the model's topics and the ground truth.
%the ground truth partition and the topics produced by a model.
%
%We are not focusing on the text-based summary of topics, as \abr{lda} generates keywords while \topicgpt{} creates phrases: the phrases are going to be more fluent but perhaps not more useful.
%
%Instead, we are going to focus on which documents are in which cluster
% partition (topic) based on a ground truth partition.
%
% \jbgcomment{Should have citations for the original metrics too}
% To measure the similarity between the ground truth partition and the
% topics produced by a model, we use several metrics:
Purity~\cite{purity}, adjusted
rand index~\cite[ARI]{ARI}, and normalized mutual information~\cite[NMI,]{NMI} have
become common alternatives to traditional coherence measures in recent topic modeling evaluations~\cite{pham2024topicgpt,li-etal-2024-improving, angelov-inkpen-2024-topic}.
%
% \abr{lda} on 85 topics, it achieves the highest purity on both datasets compared to other models. 
%
% \lorenacomment{I don't think this is the right place for it. LDA's K is a setting, and here we're talking about results} Since \abr{lda} requires predefined number of topics, we use 65 to be the topic number, which is similar to the number of gold topics from the dataset.
%
% Focusing
Relying on a single metric may introduce bias, as each can be manipulated.
For example, purity measures how well a clustering algorithm groups similar documents by comparing clusters to ground truth labels, but it can be exploited by assigning a unique label to every document.
Using all three metrics together provides a more balanced evaluation of how well the topics align with the gold topics.
%

% Since \abr{lda} has more topics than all other methods, it results in higher purity (Table~\ref{tab:cluster_metrics}).
% \lorenacomment{Why LDA generates more topics than all other methods? You can specify how many topics LDA generates... Wouldn't it be better something like ``Since the number of topics we train LDA with (equal to the number of categories) is larger than those discovered by LLoom or TopicGPT, ...''}
%
Table~\ref{tab:cluster_metrics} presents the clustering evaluation metrics for all methods. \bass{} has the highest overall scores, demonstrating that human-supervision
during the topic generation process, combined with active learning to refine the original \abr{lda} topic distribution enhances clustering performance. Notably, users only need to label about 50 documents.

\lloom{} fundamentally differs from the other algorithms, resulting in the lowest overall clustering scores.
%(Table~\ref{tab:cluster_metrics}).
%
It uses \abr{HDBSCAN}~\cite{10.1007/978-3-642-37456-2_14} to cluster \mm{}-extracted summaries, and relies on \mm{}s for topic synthesis. As a result, its topics emerge from a series of high-level summarization and data transformations, thus introducing information loss that lowers clustering metrics, even though its topics remain reasonable and appealing.

%and using \mm{} for topic synthesis, its topics are the result of a series of high-level summarization and data transformations. This process introduces information loss, which lowers clustering evaluation metrics, even though its topics remain reasonable and appealing.
%
\topicgpt{}'s performance with GPT-4 as the underlying \mm{} as underlying \mm{} matches \lda{} on \texttt{Bills}---a benchmark from its original study---but falters on \texttt{Sci-Fi}.
It generates overly generic topics, limiting its usefulness for exploring domain-specific data (\S~\ref{qualitative}; \S~\ref{app:exmaple_generated_topics}).
%\topicgpt{} appears to generate overly generic topics, showing limitedvalue to users exploring domain-specific data, as discussed further with examples in \S~\ref{qualitative}.
% We analyze this divergence from the original claim that \topicgpt{} reaches the best clustering results than \abr{lda} and BERTopic with a detailed error analysis in \S~\ref{qualitative}.
%
Performance declines further with  Mistral-7B~\cite{jiang2023mistral} and LLaMA-2-70B~\cite{touvron2023llama}.
%Additionally, we run the \topicgpt{} pipeline using  Mistral-7B~\cite{jiang2023mistral} and LLaMA-2-70B~\cite{touvron2023llama}.
%
Mistral struggles with topic generation and merging tasks, resulting in the lowest overall clustering metrics, while LLaMA-2 performs comparably to GPT-4 but still yields worse clustering metrics  (Table~\ref{tab:cluster_metrics}).
Local \mm{}s, particularly smaller \mm{}s struggle at reasoning and describing the haystack of a dataset, thus we omit using local \mm{}s to assist users exploring the documents.\footnote{Local \mm{}s are not included in the user study.}

% \zongxiacomment{Later add strength and weaknesses of topicgpt, lloom}

\subsection{Automatic Evaluation Metrics}
In data exploration, using a common tool for analyzing the same dataset improves reproducibility by ensuring consistent methodologies and conclusions~\cite{NationalAcademies2019}.
%In standard data exploration, researchers analyzing the same dataset can improve reproducibility by using a common tool that promotes consistent methodological approaches and research question analysis.
%
%Rigorous science demands reproducible methods and consistency among data exploration conclusions, as non-reproducible research lacks scientific rigor~\cite{NationalAcademies2019}.
%
To evaluate how these tools could support this data exploration to facilitate learning from data, we compare user responses across groups using two metrics: \textit{Answer consistency} and \textit{Answer quality} (\S~\ref{study_setup}).
%Thus, we evaluate the \textit{consistency} and \textit{answer quality} to compare user responses across groups (\S~\ref{study_setup}).
%
\textit{Consistency} quantifies the similarity of responses within a group, with higher similarity indicates better incorporation of corpus information.
\textit{Answer quality} measures the alignment between user responses and ideal answers.
Both metrics are computed using \textsc{all-mpnet-base-v2}.\footnote{\url{https://sbert.net/docs/sentence_transformer/pretrained_models.html}}
%

%\textit{Consistency} measures how similar and consistent among all the
%users using a tool (under the assumption that if the answers become
%more consistent, they have incorporated more information from the
%underlying corpus); \textit{answer quality} measures how similar are
%he user answers to the given ideal answers.
%
%Consistency is the cosine similarity of all answer paris within an
%experimental group (\S~\ref{study_setup}) using
%\textsc{all-mpnet-base-v2}.\footnote{\url{https://sbert.net/docs/sentence_transformer/pretrained_models.html}}
%
%\textit{answer quality} is the cosine similarity between the
%reference answer averaged over each user's answers.
%
%
% Specifically, the previous section shows that there is no significant difference of user backgrounds in the pretest stage, we aim to answer the following question: Is the pairwise answer similarity significant among groups in the post test? 
% %
% We assess whether one method leads to greater improvements in answer consistency after participants are exposed to the data.
% We compute the median transformer similarity per question and the mean across all questions for each group in Fig.~\ref{fig:automatic_evaluation}.
%
\begin{figure}[t]
    \centering
    \includegraphics[width=1.0\linewidth]{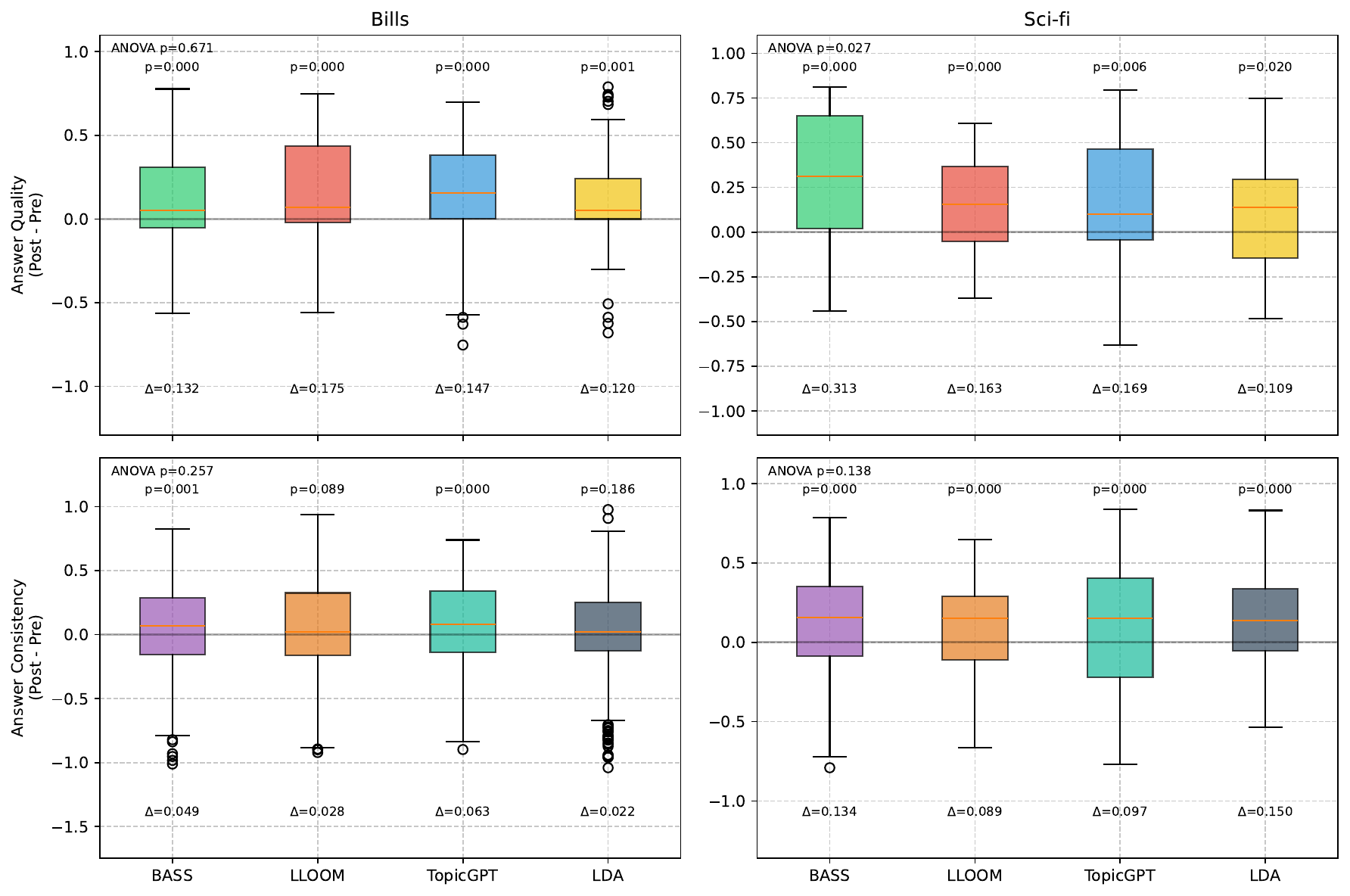}
    \caption{
    No significant differences exist among groups, except for \texttt{Sci-Fi} answer quality (one-way ANOVA, $(p < 0.05))$.  
     %There is no statistical significance among groups except\texttt{Sci-Fi} answer quality using a one way ANOVA test at significance $p$ value (0.05). 
    %The $p$ values on top of each boxplot is one-sample $t$-tests comparing each group's differences (post vs. pre) against 0, where all groups except \lloom{} on answer consistency show an increase of answer consistency and quality after users use a tool.
    The \(p\)-values above each boxplot correspond to one-sample \(t\)-tests comparing each group's pre- vs. post-task differences against 0. All groups, except \lloom{} on answer consistency, show increased answer consistency and quality after using the tool.
    Overall, \mm{}-based methods do not show significant advantages over \abr{lda} on automatic evaluations. %, but human raters \emph{prefer} responses generated by users using \mm{}-based models (Fig.~\ref{fig:bradely_terry}).
    % \zongxiacomment{P value and test statistics}
    % On the other hand, \bass{} users show the highest answer quality gain on Sci-fi datasets.
    }
    \label{fig:automatic_evaluation}
\end{figure}
% In real life datasets (\texttt{Bills}), \bass{} has the highest average topic relevance scores and answer consistency scores, while \abr{lda} is the best on synthetic \texttt{Sci-fi} data.
%
All tools help improve answers, with a statistically significant increase from pretest to posttest across groups.
Fig.~\ref{fig:automatic_evaluation} shows the results for \textit{Answer consistency} and \textit{Answer quality}.
%as the average similarity of all answer pairs per group and \textit{Answer quality} as the similarity between the reference answer and group-averaged responses. 
%with a statistically significant increase from pretest to posttest across groups.
%
Users' answers become both more consistent and better aligned with reference answers after using the tools. % (Fig.~\ref{fig:automatic_evaluation}).
However, 
%a one-way ANOVA test shows 
there are no significant difference between tools, indicating all topic models are similarly effective in helping users improve their answers in terms of {\it Consistency} and {\it Answer quality}.
%there is not a significant \emph{difference} between tools
%under a one way ANOVA test.
% that fully \mm{}-based methods are better for exploration than \abr{lda}.
%

% For both datasets per question, groups with topic models show higher average similarity scores than the baseline group in most cases.
% %
% \bass{} on average has the highest similarity per question on Bills while \topicgpt{} has the highest on News, implying that different users under the help of \mm{}s on average have more answer consistency than traditional topic models or without any tools.

\subsection{\texttt{Pairwise Response Preference}}
\begin{figure}[t]
    \centering
    \includegraphics[width=1.0\linewidth]{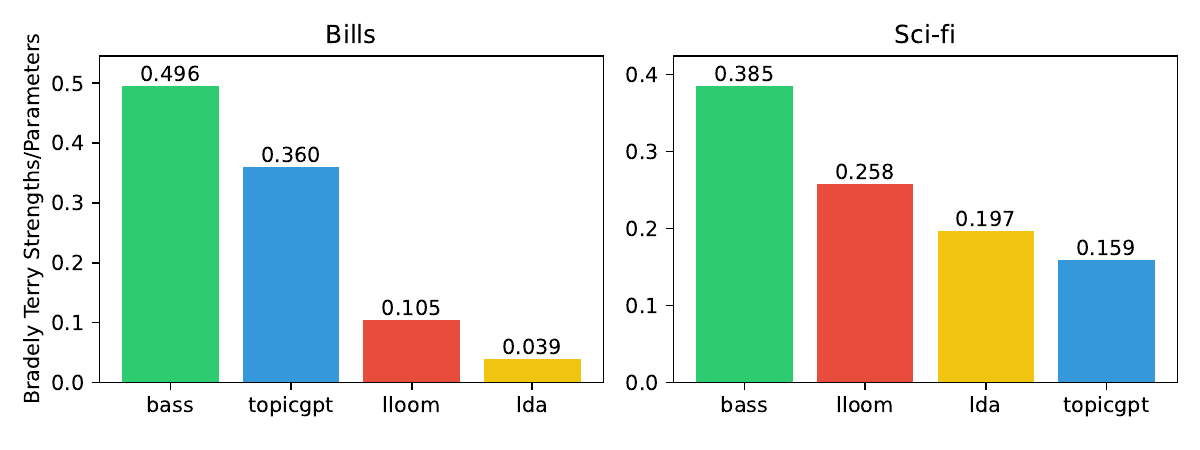}
    \caption{\bass{} has the highest preference strengths than other groups on both datasets, where users using \bass{} generate better responses than unsupervised topic models. However, the position of \topicgpt{} swaps from \texttt{Bills} to \texttt{Sci-Fi}, where \topicgpt{} only generates one super topic and three subtopics for \texttt{Sci-Fi} (\S~\ref{qualitative}).}
    \label{fig:bradely_terry}
\end{figure}

% We use the refined evaluation rubric from Table~\ref{tab:rubric} to compare the user responses generated under different conditions for the same questions.
% %
% We hire crowdsource annotators and present them with the scoring rubric, the question, the reference answer, two user responses from different groups, and the dataset to search for relevant documents by topics.
% %
% We filter annotators by an English reading related or social science background and choose their preferred response.
% %
% Specifically, the answers are randomly shuffled for each question so annotators will not know the answer groups.
% %
% We have three annotators for each dataset to annotate the data, and use the majority vote as the gold preference with a Krippendorf's alpha~\cite{castro-2017-fast-krippendorff} score of \(0.73\) for Bills and \(0.76\) for Sci-fi
% %
% We use pairwise preference with Bradley Terry Model ranking to measure the strength and likelihood users prefer responses for one topic model than the other (Fig.~\ref{fig:bradely_terry}).
Assigning a Likert score to an answer can introduce annotator bias, whereas selecting the better answer between two responses is typically a simpler and more objective task~\cite{chiang2024chatbotarenaopenplatform}.
Thus, we use the refined evaluation rubric from Table~\ref{tab:rubric} to
compare user responses generated under different conditions for the
same questions.
Prolific annotators are hired and provided with the scoring rubric,
the question, the reference answer, two user responses from different
groups, and the dataset to search for relevant documents by topic.
Annotators are filtered based on an English or Social Science
background and tasked to select their preferred response.
We randomly shuffle responses for each question so annotators cannot
identify the source groups.
%
% To ensure fairness, the responses are randomly shuffled for each question so annotators cannot identify their source groups. 
%
Each dataset is annotated by three annotators, and use the majority
vote as the gold preference, with a Krippendorff's
alpha~\cite{castro-2017-fast-krippendorff} score of \(0.73\) for \texttt{Bills}
and \(0.76\) for \texttt{Sci-Fi}.
%
% \jbgcomment{Did you cite BT the first time you used it?}
We use a Bradley-Terry Model, which measures the probability of responses from a group winning in pairwise comparisons, to rank users favoring responses of one group over another
(Fig.~\ref{fig:bradely_terry}).

\paragraph{Do \mm{}s help users generate more preferable responses than traditional topic models?} 
%
%Depending on the content of the datasets, \mm{}-based methods do not always help users generate preferable answers(Fig.~\ref{fig:bradely_terry}).
\mm{}-based methods do not always lead to preferable answers----their effectiveness depends on the dataset content (Fig.~\ref{fig:bradely_terry}).
%
%The standard \texttt{Bills} dataset is a well-organized corpus with documents that have clearly identifiable topics.
%
% \jbgcomment{Should we ask the models if they know the top-level labels in their parametric memory?}
%
%\mm{}s readily distinguish between different documents or \mm{} have strong parametric memory, and \mm{}s can effectively replace human annotators in identifying the distinct topics within the dataset.
But do \mm{}s readily distinguish between different documents, or do they rely on strong parametric memory?\footnote{We ask GPT-4 to generate a list of topics for \texttt{Bills} and alien science fiction without providing it any documents by simple prompting. Over 95\% topics GPT-4 generates fall within the \texttt{Bills} gold topics, 
while none align with the \texttt{Sci-Fi} gold topics (Appendix\ref{sec:parametric}).}

In structured, well-organized corpora with clearly identifiable topics---such as \texttt{Bills}---\mm{}s can often replace human annotators in identifying distinct topics.
%within structured datasets
%For easy datasets that need simple classification, 
In this setting, all user groups involving \mm{}s generate better responses than keyword-based topic models (\abr{lda}), achieving higher Bradley-Terry probabilities.

However, this advantage does not always hold for domain-specific datasets like \texttt{Sci-Fi}, which has abstract gold topics that \mm{}s do not easily surface.
%In contrast, the \texttt{Sci-Fi} dataset is domain-specific, with abstract god topics that \mm{}s do not easily surface.
%and the gold topics are abstract and not readily surfaced by \mm{}s: 
For example, the topic \textit{Ethics and morality: Delving into the moral and ethical dilemmas that arise from encountering a non-human intelligence...} is difficult for \mm{}s to identify.
Here, \topicgpt{} falters, \colorbox{green!15}{producing only one generic topic:} 
%\colorbox{green!15}{it only generates one generic topic} 
\textit{Science and Technology: Involves the study and application of scientific advancements.} This topic is not closely related to the corpus, and its three subtopics remain nonspecific due to \topicgpt{}'s inflexible prompt formats and pipelines. 
%nonspecific subtopics due to its  inflexible prompt formats and prompt pipelines. 
%

Moreover, unsupervised \mm{}s are sensitive to prompt templates, sometimes leading to hallucinations. For instance, \topicgpt{} \colorbox{green!15}{generated an \textit{agriculture} topic for a math dataset} (\S~\ref{app:exmaple_generated_topics}).
%
%Thus, 
\mm{} hallucination also shows that unsupervised \mm{} algorithms may be unreliable across different domain datasets.
However, when human supervision guides the generation of \mm{} topics,
\mm{} significantly improve users' ability to answer questions for both datasets (Fig.~\ref{fig:bradely_terry}).

The next section provides a detailed qualitative analysis of \bass{}, unsupervised \mm{}-based methods, and \abr{lda}, along with best practices for data exploration.

 \label{User_study}

\section{Qualitative Analysis and Best Practices} \label{sec:sections/55-qualitative} 
% \section{Qualitative Analysis}
%
% \jbgcomment{``human evaluation'' is vague, be precise}
Although pairwise response preference and Bradley-Terry scores (Fig.~\ref{fig:bradely_terry}) show that our human-supervised \mm{}-based topic model (\bass{}) has the highest winning probabilities, user comments highlight dissatisfaction with the substantial effort required, even for a simple dataset that can be fully unsupervised (\texttt{Bills}).
%
% \jbgcomment{You don't ``run'' a qualitative analysis}
%We first select user responses and comments from each model, then summarize the takeaways and things researchers should keep in mind when they have their specific goals on exploring a dataset.
We first select user responses and comments from each model, then summarize key takeaways and considerations for researchers based on their dataset goals.

\subsection{User Response Selection}
%For each group, we select user responses who left a positive comment and user responses who left a negative comment and review what users like about each tool.
For each group, we select user responses with positive and negative comments to understand what users like about each tool.
%
% \jbgcomment{You didn't use \texttt{Bills} earlier in talking about the dataset, be consistent.}
\texttt{Bills} users initially demonstrated higher uncertainty, with 38\% responding ``I don't know'' during pre-test (31\% for \texttt{Sci-Fi}). 
In post-test, some participants (\topicgpt{}: 13\%, \lloom{}: 5\%, \bass{}: 7\%) %provided answers by verbatim recitation of topic descriptions, demonstrating limited comprehension and originality.
recited topic descriptions verbatim, showing limited comprehension and originality.
% \jbgcomment{I'm not sure what this sentence means.  Is it:
% The users who saw \mm{}-generated topics relied on the summaries more
% and answered more questions with {\it ``I don't know''} or simply
% copying the summaries verbatim.  }
%
% \jbgcomment{I think the first part of this next sentence was already said, so focused on the wider variance.}
% A notable trend was the overuse of labels and copy-pasting, with a wider range of response quality, as reflected by highly varied annotator ratings. 
%A notable trend is that overuse of copying and pasting \mm{} generated labels resulting in overall higher consistency than \abr{lda} (Fig.~\ref{fig:automatic_evaluation}).
A notable trend is that excessive copying and pasting of \mm{}-generated labels leads to higher overall consistency than \abr{lda} (Fig.~\ref{fig:automatic_evaluation}).
%
% \jbgcomment{This doesn't make sense to me\dots they can't have knowledge of a dataset that doesn't exist.  Were they more willing to make stuff up, and it happened to be right?}
\texttt{Sci-Fi} is synthetically designed to ensure that documents within the same topic are interconnected and relevant to the question, requiring less domain expertise to interpret. 
As a result, \texttt{Sci-Fi} user responses often link information from multiple documents rather than copying \mm{} topic descriptions.

\paragraph{Traditional Topic Models} 
% \jbgcomment{Not a huge fan of this advantages / disadvantages structure. Would be better to talk through it in words.  But changing this is lower priority than other tasks.}
are more accessible to social scientists without advanced technical expertise, and pose fewer data privacy concerns than \mm{}s. However, their outputs are less user-friendly.
%
%\abr{lda} user reviews show that topic keywords can provide a broad overview of possible themes within a dataset, giving guidance on the types of documents to search for when answering questions.
\lda{} users' written feedbacks still show that topic keywords offer a broad overview of dataset themes, helping guide document searches when answering questions. 
%
%Additionally, the flexibility of naming topics based on keywords gives users more freedom and imagination, aiding further data analysis.
The ability to name topics based on keywords provides users with flexibility and creative freedom for further analysis.
%
%Computationally, \abr{lda} is faster and more resource-efficient compared to other approaches that prompt \mm{}s.
Additionally, \abr{lda} is computationally faster and more resource-efficient,% than \mm{}-based approaches.

However, keyword-based topics require more effort to interpret.
%
%Two users reported needing time to adapt to the tool as the initial keyword-based topics were confusing and daunting.
%
%One users mentions that \abr{lda} generates repetitive and uninterpretable topics due to its reliance on word distributions.
%
%For example, \textit{bank} and \textit{banks} can appear in the same topics, or nonsensical keywords such as \textit{thing, matter...}, can confuse and deter users.
Two users reported an initial learning curve, finding keyword-based topics confusing and overwhelming. Another noted that \lda{} generates repetitive or uninterpretable topics, such as \textit{bank} and \textit{banks} appearing in the same topic, or generic terms like \textit{thing} or \textit{matter} that confuse users.

% In summary, traditional topic models are fast, efficient, and scalable but are less user-friendly, requiring greater user effort to interpret the generated topics.

\paragraph{Unsupervised \mm{}-based methods} perform well on datasets with clear topic boundaries, generating intuitive topics that users prefer over topic names or word lists. 
%that users find more intuitive.
% On datasets with clear topic boundaries, \mm{}-based topic models
% produce topic phrases that are more favorable to users.
%
%Users prefer topic descriptions over topic name or word lists.
%
Among the models, \topicgpt{} produces more systematic topics on 
\texttt{Bills}, while \lloom{} is more stable, generalizable, and capable of extracting abstract topics across diverse datasets due to differences in prompting and topic generation algorithms (See 
%topic 
examples in \S~\ref{app:exmaple_generated_topics}).
%

%On the other hand, 
However, users find overly generic topics less useful for realistic tasks.
%, making them less useful for our realistic tasks.
%
For instance, on the \texttt{Sci-Fi} dataset, \topicgpt{} generated a single
broad topic, \textit{Science and Technology}, with three
subtopics: \textit{Non-Human Intelligence, Interspecies
Communication,} and \textit{Ethical and Moral Implications}.
In domain-specific datasets requiring abstract reasoning, \topicgpt{} tends to hallucinate,
producing overly broad or unrelated topics---e.g., {\it agriculture} for a math dataset (\S~\ref{sec:domain_specific_data}).
%and is overly broad in its topic selections, such as generating \textit{agriculture} for a math dataset, entirely unrelated to the subject (\S~\ref{sec:domain_specific_data}).
% \jbgcomment{Would be much better to have a clear example of this}
%
\lloom{} can extract abstract topics but struggles with scalability due to its summarization-based approach. 
Processing large datasets (over \(2\,000\) documents) requires chunking,
complicating topic generation. % process.
%

%Additionally, users observed inaccurate topic assignments, such
% as documents incorrectly classified under environment and policies.
Users also noted inaccurate topic assignments, with documents misclassified under broad categories like {\it Environment} and {\it Policies}.
%
%For example, we ask \textit{What policies and regulations does the U.S. government implement to address water contamination and ensure environmental protection?} in Bills.
%
%The generated topic, \textit{Environmental Efforts: Does this text relate to efforts in environmental protection or conservation?} (\lloom{}), is where users were most likely to explore relevant documents. 
For example, in the \texttt{Bills} user study, when asked \textit{What policies and regulations does the U.S government implement to address water contamination and ensure environmental protection?}, users found the \lloom{}-generated topic \textit{Environmental Efforts: Does this text relate to efforts in environmental protection or conservation?} most relevant. 
However, user feedback indicates that some documents 
%such as a proposed State Plan Amendment (SPA) 
related to Medicaid or health services are incorrectly classified under this topic. 

%
% For example, we find a document describing the process of approving or disapproving a proposed State Plan Amendment (SPA) related to Medicaid or health services, which is unrelated to environmental protection.
% \jbgcomment{again, make this explicit: what's the document, what's the topic}
%
% Unsupervised \mm{}-based models have long-context hallucinations.
%
Unsupervised \mm{}-based models are also computationally expensive, requiring long training
times, and tend to produce overly generic topics in domain-specific datasets. 
% (\S~\ref{appendix} provides additional examples).
%for more details on the topics generated for two additional domain-specific datasets not included in the user study).

\begin{table*}[h!]
\centering
\small
\begin{tabular}{>{\centering\arraybackslash}m{0.15\textwidth} m{0.25\textwidth} m{0.25\textwidth} m{0.25\textwidth}}
\toprule
\textbf{Approach} & \textbf{Advantages} & \textbf{Disadvantages} & \textbf{Suitable For} \\
\midrule
Traditional Topic Models &
\begin{itemize}[leftmargin=*,noitemsep,topsep=0pt]
\item Fast computation
\item Low resource use
\item Less data privacy concerns
\item Broad theme overview
\end{itemize} &
\begin{itemize}[leftmargin=*,noitemsep,topsep=0pt]
\item Less user friendly
\item Potential repetitive keywords, topics, and useless topics
\end{itemize} &
\begin{itemize}[leftmargin=*,noitemsep,topsep=0pt]
\item Large document collections
\item Low resource Settings
\item Preliminary exploratory analysis
\end{itemize} \\
\midrule
Unsupervised \mm{}-based Models &
\begin{itemize}[leftmargin=*,noitemsep,topsep=0pt]
\item Descriptive topic phrases and descriptions
\item Sometimes can induce abstract topics
\item Cluster based on semantic, not words distribution
\end{itemize} &
\begin{itemize}[leftmargin=*,noitemsep,topsep=0pt]
\item Overly generic topics
\item Document assignment hallucinations
\item Expensive computation
\end{itemize} &
\begin{itemize}[leftmargin=*,noitemsep,topsep=0pt]
\item Small document collections
\item Data with clear topic boundaries
\end{itemize} \\
\midrule
Supervised \mm{}-based Models &
\begin{itemize}[leftmargin=*,noitemsep,topsep=0pt]
\item No need to traverse all documents manually
\item Flexible user topic definition and supervision
\item Avoids garbage topics
\end{itemize} &
\begin{itemize}[leftmargin=*,noitemsep,topsep=0pt]
\item Need mental efforts
\item Require user expertise
\item Inconsistent effectiveness
\end{itemize} &
\begin{itemize}[leftmargin=*,noitemsep,topsep=0pt]
\item Iterative and advanced data analysis
\item Answering abstract conceptual questions
\item Low resource settings
\end{itemize} \\
\bottomrule
\end{tabular}
\caption{While \mm{}s can generate more interpretable topics than traditional topic models, they face challenges in scaling to large corpora. Combining human input with \mm{}s can help reduce hallucinations and improve scalability.
%Combining humans with \mm{}s can reduce \mm{} hallucinations and brings scalability.
}
\label{tab:topic_modeling_comparison}
\end{table*}

\paragraph{Human supervised \mm{} topic generation (\bass{})} gives users full control over topic definitions, allowing them to define and refine topics to fit their needs.
Unlike traditional topic models, which can generate uninterpretable
topics, or unsupervised \mm{} models, which may hallucinate
topics, supervised models ensure user-defined topics. %have a users control topic definitions.
Users appreciate \bass{} for allowing them to view \mm{}-generated topic
suggestions while also revising or defining their own topics. % to complete their tasks.
The active learning process reduces the necessary number of queries to the \mm{}, lowering prompting costs compared to unsupervised \mm{}-based approaches (\S~\ref{sec:cost}).

%A disadvantage of human supervision is that users report difficulties in determining when to stop
%the generation process, as the criteria for stopping the process is
%not defined.
A drawback of the human supervision loop in \bass{} is that users struggle to determine when to stop the generation process, as stopping criteria are undefined.
Reviewing documents with \mm{}-generated suggestions can be
overwhelming, as users must read through more documents than
unsupervised methods, even with \mm{} assistance.
On simpler datasets like \texttt{Bills}, user supervision offers limited
benefits, with users approving suggested topics 93\% of the time
compared to 62\% on the \texttt{Sci-Fi} dataset.
In such cases, 
%relying on 
unsupervised \mm{}-based models can significantly reduce user effort.

% In conclusion, supervised \mm{}-based topic models provide a balance between automation and user control, though they can be resource-intensive and challenging to manage for large or complex datasets.

% \jbgcomment{THis section promised qualitative analysis, but it starts with conclusions.  Save that for later.}

\paragraph{Are \mm{}-based approaches ready to replace traditional topic models for data exploration?}
Corpora exploration and understanding is an iterative process and is rarely a matter of
running an algorithm once and retrieving your desired answers.
While emerging \mm{}-based approaches are exciting and new, there are
important considerations to be made to use them suitably: Are the predominant
metrics reliable, and what are reliable ways to evaluate emerging \mm{}-based methods?
Traditional topic models use automatic topic coherence, but this metric does not generalize to neural topic models~\cite{hoyle-21}.
% does not always reflect good and interpretable outputs for neural
% models.
%~\cite{aletras-stevenson-2013-evaluating, coherence}.
% \jbgcomment{some of these have been cited elsewhere, tone down self-cites}
%
%the emergence of new variations, such as
%prompt-based \mm{} topic models, makes it increasingly challenging
%to evaluate their relative performance.
Additionally, coherence is not applicable to \mm{}-based methods, making it difficult to evaluate their performance and usefulness.
Trade-offs between approaches are not always clear: \mm{}-based
methods are more computationally expensive and generate overly generic topics.
%
%While traditional \abr{lda} has overall higher clustering metrics than
%unsupervised topic models (Table~\ref{tab:cluster_metrics}), automatic metrics
%(Fig.~\ref{fig:automatic_evaluation}) and human evaluations
%(Fig.~\ref{fig:bradely_terry}) show that \mm{}-based approaches are
%not always the best for every task, and \abr{lda} might still be more useful than fully unsupervised \mm{} approaches.

While traditional \lda{} outperforms unsupervised \mm{}-based methods (Table~\ref{tab:cluster_metrics}), automatic metrics (Fig.~\ref{fig:automatic_evaluation}) and human evaluations (Fig.~\ref{fig:bradely_terry}) indicate that \mm{}-based methods are not always the best for every task, and \abr{lda} may still be more useful than fully unsupervised \mm{} approaches.

All tools have their trade-offs and user preferences.
In our analysis, \mm{}s are still not a replacement for traditional topic models.
Further improvements in \mm{}-based methods for large corpus understanding (especially multi modal corpuses that contain charts, plots)---such as, reducing hallucinations, improving scalability, minimizing human intervention, aligning topics with user intents, and lowering costs--- is necessary to drive their adoption and accessibility in the social science domain~\cite{Huang_2025, vlmSurvey}.
 \label{qualitative}

\section{Conclusion} \label{sec:sections/90-conclusion} % The type of topic model selected must align with the nuanced needs and requirements of the dataset and research objectives at hand. A model that performs well on one dataset or task may not necessarily be the optimal choice for a different context. Researchers should assess the strengths and limitations of various topic modeling techniques in the context of their own work, rather than making assumptions based on generic performance claims.

While advocates push for \mm{}-based solutions for data exploration, their
evaluation remains limited to cluster-based assessment or topic matching against gold labels.
Our more realistic evaluation shows that while all topic models can help humans understand a dataset, \mm{}-based approaches still have their limitations: instability, hallucination, scalability, and inflexibility.
Traditional topic models are still the cheapest option for users conducting preliminary and simple data exploration.
That said, our results confirm that people like \mm{} outputs.

Our proposed model, \bass{}, helps address some of these concerns
for trickier datasets---it is cheaper (\S~\ref{sec:cost}) and less vulnerable (\S~\ref{tab:topic_generation_comparison}) to
hallucination.
However, the cognitive effort required can be excessive for simple datasets, 
making \bass{} best suited  
%and it is best recommended 
for advanced data exploration or highly motivated users.
Nonetheless, users often prefer the topics they build
themselves~\cite{NORTON2012453}.
In sum, there is no definite answer which topic model is best for all circumstances, nor can \mm{}-based methods fully replace
traditional or human-supervised topic models.
Future work should aim to reduce the cost of \mm{}-based methods by developing stronger local models capable of systematically comprehending large document corpora, generating more interpretable topics, and reducing hallucinations without human supervision. Additionally, efforts should focus on grounding model outputs to the source documents and minimizing human effort by improving \mm{}s’ reasoning across multiple documents.
% Future work should focus on reducing hallucinations 
%reasoning abilities spanning multiple documents.

% The choice of what tools to use relies on the user goals,
% applications, and resource availabilities to maximize the advantages
% of a tool (Table~\ref{tab:topic_modeling_comparison}).

% \jbgcomment{Points the conclusion needs to hit:

% \url{https://myscp.onlinelibrary.wiley.com/doi/10.1016/j.jcps.2011.08.002}

% \begin{itemize*}
%     \item Most topic overviews help, and users can also do well when they have limited tools
%     \item Users prefer using tools where they have more of a say in how the labels are formed
%     \item This up front investment pays off by shorter time to answer questions (in our contrived example)
%     \item future work should further reduce the cost by using cheaper models, more directly connect the models for labeling and topic distributions, and provide more flexibility for users
% \end{itemize*}

% }

% \jbgcomment{Need bst to cut after 10 authors, share with everyone else}
 \label{conclusion}

\section{Limitations} \label{sec:sections/110-limitations} Data exploration is a complex task that relies heavily on human reasoning and analysis to produce meaningful insights. 
While manually sorting data is cumbersome and inefficient, tools like topic models can significantly make the process more efficient and effective for social scientists.
With the growing usage of \mm{}s in social science tasks, variations of topic modeling methods are shifting to \mm{}-based~\cite{pham2024topicgpt, lam2024concept}, which also raises important questions and challenges about their effectiveness, accuracy, and interpretability~\cite{li-etal-2024-pedants, li2025benchmarkevaluationsapplicationschallenges, hoyle2021automated,doogan-buntine-2021-topic, mondal2024scidoc2diagrammermafgenerationscientificdiagrams}.
To address these concerns, we evaluate and compare traditional topic models with \mm{}-based approaches using a combination of metrics: clustering evaluation, automatic metrics for pretest and post tests. 
Additionally, we use human pairwise answer preference, analyzed through the Bradley-Terry model~\cite{Bradley1952RankAO}, to incorporate a social science-inspired application perspective rather than relying solely on automated evaluations. 
Although this evaluation method provides valuable insights into the capabilities of \mm{}-based tools for data exploration applications, it is challenging to scale~\cite{zhou2024multi}.
In addition, despite these advances of \mm{}s for data exploration, no current approach can fully balance topic interpretability, user-friendliness, hallucination, and minimal user input. 
While \bass{} appears to help users generate effective responses and topics the most, it requires significant human supervision, particularly on simple datasets that can be fully automated.
Future work could focus on developing more user-friendly methods to reduce the need for extensive human supervision in topic generation. 
A promising direction is a hybrid approach: leveraging traditional clustering techniques to generate initial clusters, using \mm{}s to produce topics, and using a confidence detector to identify problematic topics for user correction. 
This approach can minimize user effort by eliminating the need to start the topic generation from scratch while also reducing the cost of excessive \mm{} prompting.
 \label{limitation}

\section{Ethics} \label{sec:sections/120-ethics} We received approval from the Institutional Review Board before initiating the user study. 
All participants are based in the United States or United Kingdom. 
Users are required to review an instruction and consent statement before participation commitment. 
They have the option to withdraw if they disagree with the terms. Throughout the study, no personal information that could reveal identities is collected. To the best of our knowledge, our study presents no known risks.
Non-identifiable personal information is collected throughout the study.
The compensation base rate is
$\$6.5$, which could increase to $\$17$ per hour if their answers are
deemed unlikely to be AI-generated and showing they have done the
task. 
We use a fine-tuned \textsc{RoBERTa}~\cite{sivesind_2023, zhou2024teachingassistantintheloopimprovingknowledgedistillation} to
reject users likely submitting AI-generated responses.
%
% They are provided with task instructions at the beginning
% (Appendix~\ref{tab:congressional-bills-study},~\ref{tab:news-post-analysis})
% and can quit the study at any time.
Users are notified that they can quit the study at any time for any personal reasons.
Upon completion, users are prompted with two survey questions to rate
their experience using the tool (Appendix~\ref{fig:survey_ratings}).
 \label{ethics}

\section{Acknowledgements} \label{sec:sections/130-acknowledgements} % We thank anonymous reviewers and Alexander Hoyle and Kyle Seelman for their insightful comments for helping us make our paper experiments and arguments more solid. We thank Emily Walpole and Juan Fung's community resilience groups for taking their time participating our expert verification experiment and providing valuable qualitative comments and feedback. Zongxia Li, Andrew Mao, Daniel Stephens, and Pranav Goel's contributions are supported by the \abr{nist} Professional Research Experience Program. 

We thank anonymous reviewers for providing insightful comments for helping us make our paper experiments and arguments more solid. 
Zongxia Li and Daniel Stephens' contributions are supported by the \abr{nist} Professional Research Experience Program. 
The work of Lorena Calvo-Bartolomé has been partially supported by Grant PID2023-146684NB-I00 funded by MICIU/AEI/ 10.13039/501100011033 and by ERDF/UE.
 \label{acknowledgements}

% \bibliography{bib/journal-full,bib/jbg,bib/zongxia, bib/andrew}
\bibliography{bib/custom}
\bibliographystyle{style/acl_natbib}

\section{Appendix} \label{sec:sections/200-appendix} % \newpage
\appendix
\begin{figure*}[!t]
    \centering
    \includegraphics[scale=0.44]{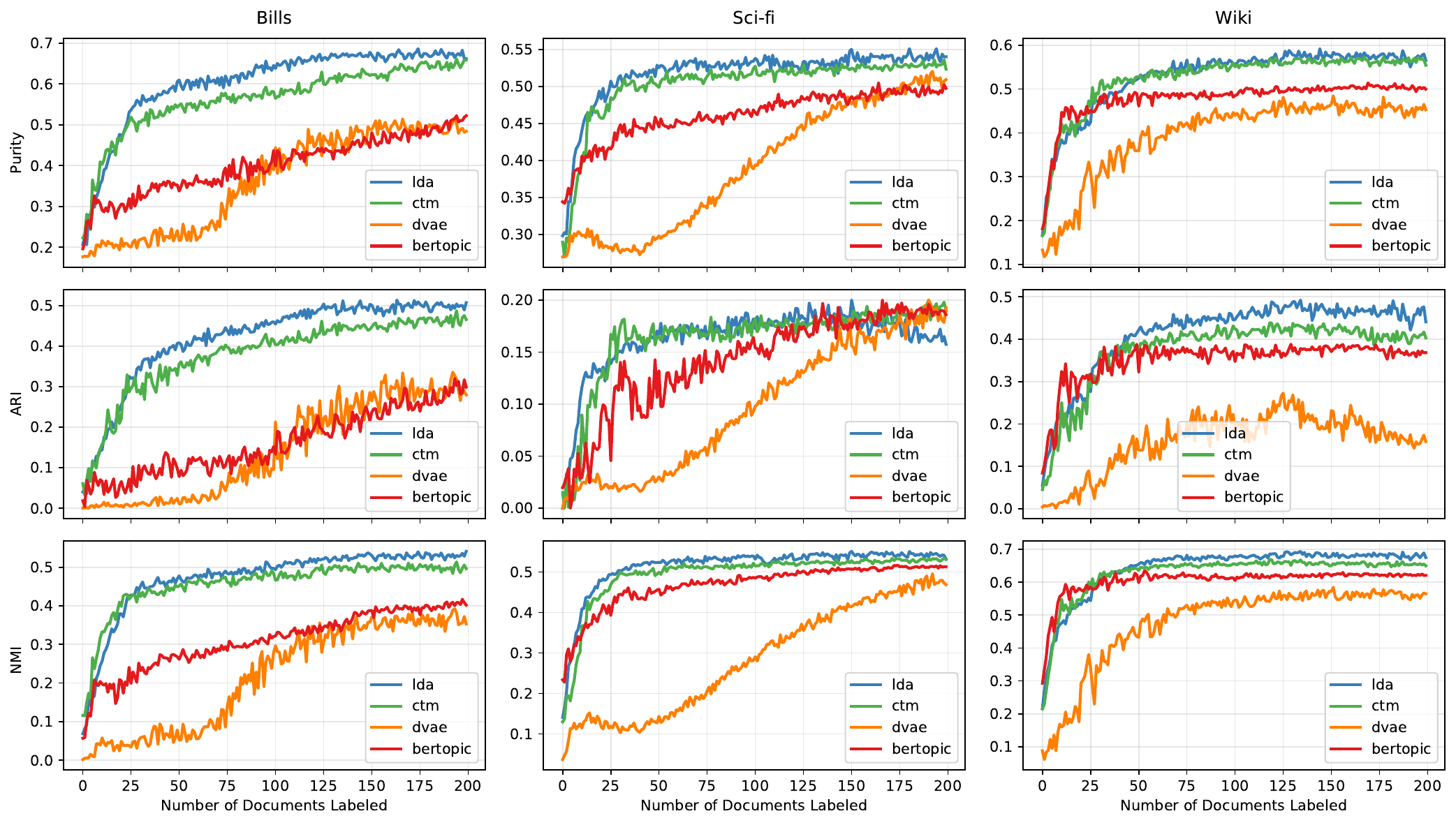}
    \caption{
    Synthetic user study experiment comparison of {\color{bertopiccolor}\bertopic{}}, {\color{ldacolor}\lda{}-\mallet{}}, {\color{ctmcolor}\ctm{}} and {\color{dvaecolor}\dvae{}} across the datasets of \texttt{Bills}, \texttt{Sci-Fi}, and \texttt{Wiki}. Each row represents a different label-centric clustering metric: Purity (top), Adjusted Rank Index (ARI, middle), and Normalized Mutual Information (NMI, bottom). The x-axis shows the number of labeled documents, while the y-axis the respective metric scores. {\color{ldacolor}\lda{}-\mallet{}} and {\color{ctmcolor}\ctm{}} generally achieve better clustering metrics with the same number of documents labeled, while {\color{bertopiccolor}\bertopic{}} lags behind in most cases. {\color{dvaecolor}\dvae{}} exhibits more variability and lower clustering scores across datasets. However, {\color{ldacolor}\lda{}-\mallet{}} is still considered the best compared to other neural topic models.
    }
    \label{fig:synthetic_experiment}
\end{figure*}

\section{Synthetic Experiments}\label{appen:synthetic_experiments}
Traditional topic models have many variants, including Bayesian probabilistic approaches (BTM) (e.g., \lda{}), neural methods (NTM), including contextualized topic models~\cite[\ctm{},]{bianchi2020pre,kitty_CTM} and the Dirichlet Variational Autoencoder Model~\cite[\dvae{}]{burkhardt2019decoupling}, as well as clustering-based models like \bertopic{}~\cite{bertopicMark}.

Our purpose in this work was to evaluate whether the new paradigm of \mm{}-based topic models truly surpasses traditional models for learning about data. Given the cost of human-based evaluation, which this work heavily relies on, testing all state-of-the-art traditional models is impractical. Instead, we select {\bf one traditional variant} and compare it against the {\bf two main \mm{}-based approaches}, \topicgpt{} and \lloom{}, as well as {\bf our supervised \mm{} model, \bass{}}. This section details the experiments conducted to select \lda{} as representative of the traditional variant.

Neural topic models generally achieve higher coherence scores than \lda{}, but~\citet{hoyle2021automated, doogan-buntine-2021-topic, li-etal-2024-improving} show that traditional coherence metrics do not generalize well to neural topic models,  as they tend to favor NTM topics over BTM ones without fully correlating with human assessments.
Hence, due to the lack of reliable automatic evaluation metrics for both Bayesian and neural topic models, we mimic~\citet{li-etal-2024-improving}'s synthetic experiment to find the most suitable traditional topic model to use in our user study.

We first provide a brief description of the evaluated topic models and their configurations, along with the datasets used for evaluation. We then detail the synthetic experiment process and results, demonstrating that \mallet{}-\lda{} is the most suitable traditional topic model for our user study.

\subsection{Datasets}
As datasets, we use \texttt{Bills} and \texttt{Sci-Fi}, as defined in \S~\ref{sec:datasets}, along with an additional dataset, \texttt{Wiki}, sourced from Wikipedia articles~\cite{Merity2017RegularizingAO}. The \texttt{Wiki} dataset consists of \num{14,290} articles spanning 15 high-level and 45 mid-level topics, including widely recognized public topics such as music and anime. It serves as a traditional baseline for topic modeling evaluation.

We chose not to include this dataset in the main experiments because it is part of \mm{}'s pretraining data. Additionally, most Wikipedia topics are widely known and highly diverse, making it difficult to measure knowledge gain according to the definition used in this work. 

For the candidate models representing the traditional variant of topic modeling, we consider \mallet{}-\lda{}, as explained in the main paper (see \S~\ref{subsec:study_conditions}), along with the following models\footnote{All implementations are integrated into our topic modeling training class\footnote{\url{https://github.com/zli12321/TopicModelLLM/blob/main/src/topic_modeling/topic_model.py}}.}:

\paragraph{\ctm{}.} Specifically, its \abr{CombinedTm}~\cite{kitty_CTM} variant extends \abr{Prodlda}~\cite{prodLDA} by incorporating Sentence-\abr{bert} embeddings~\cite[\abr{sbert}]{Reimers2019SentenceBERTSE} into the \abr{bow} representation used as input for its encoder-decoder architecture. The inference network transforms these combined representations into continuous latent document representations, while the decoder reconstructs the \abr{bow}. We use the authors' original implementation\footnote{\url{https://github.com/MilaNLProc/contextualized-topic-models}}, keeping all settings at their default values.

\paragraph{\bertopic{}.} This model follows an engineering-driven approach, generating topics by clustering \abr{sbert} embeddings without relying on word-topic or document-topic distributions. These distributions are approximated post hoc after clustering and dimensionality reduction. We train the model with \texttt{calculate\_probabilities=True}, ensuring that topic probabilities are computed for each document during the HDBSCAN clustering step. All other parameters remain at their default values, and we use the original implementation of the author\footnote{\url{https://github.com/MaartenGr/BERTopic/tree/master}}.

\paragraph{\dvae{}.} \citet{burkhardt2019decoupling} proposed reparameterizing the Dirichlet prior using Rejection Sampling Variational Inference (RSVI), preserving the properties of \lda{}-based methods while balancing interpretability and likelihood optimization. We utilize the implementation from \citet{hoyle2021automated}\footnote{\url{https://github.com/ahoho/dvae}}, where RSVI is replaced by path-wise gradients~\cite{jankowiak2018pathwise}. We set the number of training iterations to 250, while all other parameters remain at their default values.

Note that all of the latter are neural-based topic modeling algorithms. We do not consider other Bayesian-based topic models, as the superiority of \lda{}, particularly in its \mallet{} implementation, has been demonstrated multiple times in the literature~\cite{hoyle-21,doogan-buntine-2021-topic, an2023sodapopopenendeddiscoverysocial, liu2024understandingincontextlearningcontrastive}.
%

%\subsection{Synthetic Experiment Setup}
%
%Neural topic models generally achieve higher coherence scores than \abr{lda}, but \citet{hoyle2021automated, doogan-buntine-2021-topic, li-etal-2024-improving} show that traditional metrics do not generalize to neural topic models, where neural topic models often generate less user-interpretable and favored topics compared to \lda{}. 
%
%Due to the lack of reliable automatic evaluation metrics for topic models, we adopt \citet{li-etal-2024-improving}'s synthetic experiment pipeline to find the most suitable traditional topic models to use in our user study to reduce unnecessary human study costs. 
%

%\paragraph{Algorithmic definitions.} We train four topic models - \mallet{}-\lda{}, BERTopic, \ctm{}, and \dvae{} - on the Bills, Sci-fi, and Wiki, each configured to generate 65 topics.
%

\subsection{Synthetic Experiment Setup}
\label{app:synthetic_experiment_setup}
We train one topic model per model type---\mallet{}-\lda{}, \bertopic{}, \ctm{}, and \dvae{}---and dataset---\texttt{Bills}, \texttt{Sci-Fi}, and \texttt{Wiki}, with 65 topics. 

Let \(k^{*}\) be the index of the most predominant topic for document \(d\), and let \(\theta^d_{k^*}\) be its corresponding probability, where:
\begin{equation}
    k^{*} = \arg \max_{i=1}^{K} \theta_i^d
\end{equation}

Let \(L\) be the label set probability distribution for document \(d\), we train a logistic regression active learning classifier with \(\theta^d_{k^*}\) as feature and compute the classifier entropy as:

\begin{equation}
\mathbb{H}_{d}(L) = \mathbb{H}_d(L) \cdot \theta^d_{k^*},
\label{eq:topic_document_preference}
\end{equation}

%\begin{equation}
%\theta^d_{\text{max}} = \max_{i=1}^{K} \theta^d_i.
%\label{eq:topic_probability}
%\end{equation}

%For any document with a topic distribution represented by a topic model as \( \theta^d \equiv \{\theta^d_1, \theta^d_2, \ldots, \theta^d_K\} \), its predominant topic
%\begin{equation}
%\theta^d_{\text{max}} = \max_{i=1}^{K} \theta^d_i.
%\label{eq:topic_probability}
%\end{equation}

%We train a logistic regression classifier with topic model vector representations as features and compute the classifier's entropy:
%\begin{equation}
%\mathbb{H}_{d}^{t}(L) = \mathbb{H}_d(L) \cdot \theta^d_{\text{max}},
%\label{eq:topic_document_preference}
%\end{equation}
where higher entropy indicates greater classifier uncertainty in document classification.
In the original study by~\citet{li-etal-2024-improving}, a user-in-the-loop approach is employed, wherein a user iteratively revises the documents suggested by the active learning classifier. Here, we approximate user annotations by utilizing the dataset' gold labels as pseudo-labels. Since these labels are carefully curated, they provide a reliable approximation of human annotations and may even offer a slight advantage, as human annotators tend to assign more specific labels, intentionally introducing additional variability.

Our document selection process follows two steps: first we identify the topic \(k_H\) that shows the highest median document entropy among all \(K\) topics, indicating the topic where the classifier shows the most uncertainty. 
Then we choose the document with the highest entropy within topic \(k_H\) as the next document as the next document to be labeled and update the classifier using its corresponding pseudo-label.

%We use gold labels as pseudo-labels and iteratively select documents for labeling to simulate user input from our user study.
%
We use incremental learning~\cite{incrementallearning} to train and update the logistic regression classifier and compute the purity, \abr{ari}, and \abr{nmi}.
We retrain the classifier if a new label class if introduced.
For each topic model, we perform five iterations of simulated user labeling, labeling up to \textit{200} documents per iteration---the maximum number of documents that could be labeled by users within an hour. We then compute the median value for each document within each group.

\subsection{Results}
Fig.~\ref{appen:synthetic_experiments} shows the label-centric clustering metrics obtained as a result of the synthetic experiment for each dataset and topic model.

\paragraph{Mallet \abr{lda} outperforms other neural topic models on three clustering metrics on all datasets with equal number of documents labeled (Fig.~\ref{fig:synthetic_experiment}).}
\abr{ctm} is the only neural topic model that achieves comparable to \abr{lda} clustering performance. Thus, we choose \abr{lda} as the most suitable traditional topic model in our real user study.

\section{Time and monetary cost}
\label{app:time_monetary_cost}
Table~\ref{sec:cost} shows the average amount of time and costs to train each model on a size 10,000 dataset (Bills). 
\abr{lda} is the cheapest and fastest than any other models.
The other fully automated \mm{} approaches, however, are more expensive than adding human-in-the-loop approach (\bass{}).
Adding human-in-the-loop for \mm{}-aided data exploration can be cheaper and more efficient than existing fully \mm{}-based approaches.

\label{sec:cost}
\begin{table}[h!]
\small
\centering
\begin{tabular}{ccc}
\hline
% \rowcolor{gray!50}
{\bf Method} & {\bf Train Time} &  {\bf Cost}  \\ \hline
\lda{} & 5 mins & Free \\ 
\lloom{} & 30 mins & \$40 \\ 
\topicgpt{} & 9 hrs & \$65 \\
\bass{} & User dependent: 1 hr & \$30 \\
\hline
\end{tabular}
\caption{The train time and cost for each method is an approximation. Specifically, for all models besides, we use GPT-4o as the prompt model to generate topics. The estimated cost of \bass{} is one user hour cost \$20 plus the expected prompt cost \$10.}
\label{tab:cost}
\end{table}

% \section{User Study Page}
% \label{sec:user_study_section}
% See Fig.~\ref{fig:bass_topic_generation} for user study interface for reference.

% \begin{figure*}[!t]
%     \centering
%     \includegraphics[scale=0.8]{figures/demo.pdf}
%     \caption{
%     % Topic-generation process for \bass{} users. From top to bottom, and left to right: users are presented with a document and its summary (automatically generated by an \mm{}), three \mm{}-suggested labels (from which they can select one or input a new one), the already generated topics (hover to see the number of assigned documents), and a search bar to find similar documents using keyword and TF-IDF search.
%     Users will be displayed with selected documents and topics generated by topic models or topics they have added (hover to see the number of assigned documents). In \bass{}, the UI includes three \mm{}-suggested labels (from which they can select one or input a new one), but the users start with empty set of topics. The search bar can find similar documents using keyword and \textsc{tf-idf} search. Users in the baseline group will not have anything \mm{} suggested labels nor topics generated by a topic model, but they can add and make new labels.
%     }
%     \label{fig:bass_topic_generation}
% \end{figure*}

\section{Generated questions and Evaluation Rubric}
Two expert social scientists design a rubric (Table~\ref{tab:questions}) to evaluate the quality of user responses. 
%
%See Table~\ref{tab:questions} for reference.

\label{appendix:rubric}
\begin{table*}[h]
    \centering
    \tiny
    \renewcommand{\arraystretch}{1.2}
    \begin{tabular}{p{0.025\textwidth}p{0.425\textwidth}|p{0.025\textwidth}p{0.425\textwidth}}
    \toprule
    \multicolumn{2}{c}{{\bf Bills}} & \multicolumn{2}{c}{{\bf Synthetic Science Fiction}} \\
    \midrule
    1 &  What policies and regulations does the U.S. government implement to address water contamination and ensure environmental protection? & 1 & How did human perceptions of identity change when confronted with the non-human intelligence? \\
    2 & What are common policy actions taken by governments in the United States to manage land use effectively? & 2 & How did the non-human intelligence perceive humanity during its interactions? \\
    3 & Which demographic or age groups are targeted by government initiatives to enhance educational opportunities and benefits? & 3 & What challenges did the humans face when trying to communicate with the non-human intelligence? \\
    4 & What basic rights should people have when receiving care at home? & 4 & What were the consequences of the successful communication with the non-human intelligence? \\
    5 & What do policies about land use and wildlife have in common? & \multicolumn{2}{c}{} \\
    6 & Why does the government sometimes pause taxes on importing certain chemicals and materials? & \multicolumn{2}{c}{} \\
    \bottomrule
    \end{tabular}
    \caption{Pre-test and post-test questions for both datasets. The test questions are testing the users' understanding of topics in the dataset, not testing users' ability to find a specific document to find the answer.}
    \label{tab:questions}
\end{table*}

\begin{table*}[h]
    \centering
    \small
    \renewcommand{\arraystretch}{1.2}
    \begin{tabular}{c|p{0.4\textwidth}|p{0.4\textwidth}}
    \toprule
    \textbf{Score} & \textbf{Judgment} & \textbf{Examples} \\
    \midrule
    1 & Very low quality: The response is irrelevant to the subject, showing no understanding or effort. & 
    \begin{itemize}[leftmargin=*]
        \item Blank response
        \item "Affordable care"
        \item "Love"
        \item "I don't know"
        \item "no idea"
    \end{itemize} \\
    \midrule
    2 & Low quality:
    \begin{enumerate}[leftmargin=*]
        \item Shows minimal relevance or understanding of the subject, with little or limited effort.
        \item Using just the topics to answer the questions without providing explanation.
    \end{enumerate} & 
    \begin{itemize}[leftmargin=*]
        \item "trade, tariffs…"
        \item "Alien communication"
        \item "This document discusses…" (direct copy of the summary or label)
        \item Ballast Water Management Act (BWMA)
    \end{itemize} \\
    \midrule
    3 & Fair quality:
    \begin{enumerate}[leftmargin=*]
        \item Shows basic understanding of the subject and the dataset, showing some effort akin to a lay person's perspective.
        \item The response answers the question based on a single document, not the theme across multiple documents.
    \end{enumerate} & 
    \begin{itemize}[leftmargin=*]
        \item "Clean Water Act (CWA): This law regulates discharges of pollutants into U.S. waters and sets water quality standards for surface waters."
        \item young people, school leaver age 16-18 and young adult 18-26
    \end{itemize} \\
    \midrule
    4 & High quality:
    \begin{enumerate}[leftmargin=*]
        \item Shows good understanding of the subject, suggesting above-average knowledge or effort.
        \item The answers are from the documents. However, just a list of related documents or titles for the question are provided. Little analysis or insights, synthesis of those documents are given.
    \end{enumerate} & 
    \begin{itemize}[leftmargin=*]
        \item "The US government enforces Clean Water Act, regulating pollutants in waterways, Safe Drinking Water Act, ensuring safe public drinking water. EPA monitors."
        \item Younger age groups, likely from age of 4 or 5 up to 18.
        \item Also perhaps training programs are targeted at the unemployed.
    \end{itemize} \\
    \midrule
    5 & Very high quality:
    \begin{enumerate}[leftmargin=*]
        \item Shows exceptional understanding of the subject, indicating expertise or extensive effort.
        \item The answers are across themes that cover multiple documents.
        \item The answers are a synthesis, reasoning, and analysis of contents from multiple documents.
    \end{enumerate} & 
    \begin{itemize}[leftmargin=*]
        \item "Government programs frequently focus on providing assistance to low-income families, students in under-resourced schools, individuals who are the first in their family to attend college, and adults looking to enhance their job skills through training programs. Furthermore, certain programs may target minority communities and people with disabilities."
        \item "Home care recipients deserve dignity, respect, privacy, informed choices, tailored care, safety, and autonomy for well-being."
    \end{itemize} \\
    \bottomrule
    \end{tabular}
    \caption{Evaluation Scoring Rubric for Response Quality. We rate answers based on the refined rubric to reduce individual annotator subjectivity and biases.}
    \label{tab:rubric}
\end{table*}

% \begin{table}[t] % This specifies that the table should appear at the top of the page
% \centering
% \tiny
% \renewcommand{\arraystretch}{1.5} % Increase row spacing
% \begin{tabular}{p{0.5cm}p{6.3cm}} % Adjust the column widths to fit one column of the page
% \hline
% \textbf{Score} & \textbf{Criteria} \\ \hline
% $1$ & Contains I don’t know. Suspicious of AI generated text-- possibly very long and overly detailed answers. Answer is not relevant to the question or not informative; answer is vague. \\ 
% $2$ &  The answer uses outside knowledge other than the knowledge from the dataset. Answer is just a topic name or a list of topic keywords generated from the models.\\ 
% $3$ & The answer uses some document knowledge (maybe partly from the documents in pretests and answer the questions without having access to the documents), but it is too generic to provide an answer to the question. The supposed answer is just a copy-paste of a relevant document/topic description. \\
% $4$ &  The answer is mostly from the dataset documents but only uses knowledge of a very specific document or passage to answer the question without connecting information for multiple documents. \\
% $5$ & The answer is built based on multiple documents from the dataset and answers the question. \\
% \hline
% \end{tabular}
% \caption{We give an answer rating based on the rubric to reduce individual annotator subjectivity and biases.}
% % \jbgcomment{Use active verb "we rate answers"}
% \label{tab:rubric}
% \end{table}

\section{Parametric Memory and Generated Topics}
\label{sec:parametric}
A strong parametric memory of the topics in the datasets can affect the generated topic outputs. 
We examine GPT-4 parametric memory on the two test sets without providing any additional information about the documents, etc.
Unsurprisingly, GPT-4 generates 20 topics for  Bills and almost all of them are similar or overlap with the gold topics: \textit{agriculture, health care, education, environment and conservation, defense and national security, taxation and revenue, veteran affairs, energy and utilities}...
GPT-4 generates two topics that are similar to that in the Sci-fi data: \textit{first contact protocols}, \textit{Utopian or Dystopian Alien Societies}.
The remaining topics are not relevant to the gold topics in the dataset.

\mm{} API costs and time investments are important considerations for most users. Table \ref{tab:cost} summarizes the time and monetary costs for each method. For the automatic control groups, the time refers to the training time for topic generation, which depends on the number of documents in the dataset. In contrast, for \bass{}, time depends on the user's knowledge of the dataset and the diversity of topics: the more varied the topics and the less familiar the user is with the dataset, the more documents they need to review, leading to higher costs. Results show that \lda{} is by far the most efficient and cost-effective method among the four evaluated.

% \ahintext{If you're looking for places to cut, I would put this in an appendix, or summarize it in a sentence (rather than taking up space with a table)}

\begin{figure}[t]
    \centering
    \includegraphics[trim={0cm 0cm 0cm 0cm}, clip, width=0.48\textwidth]{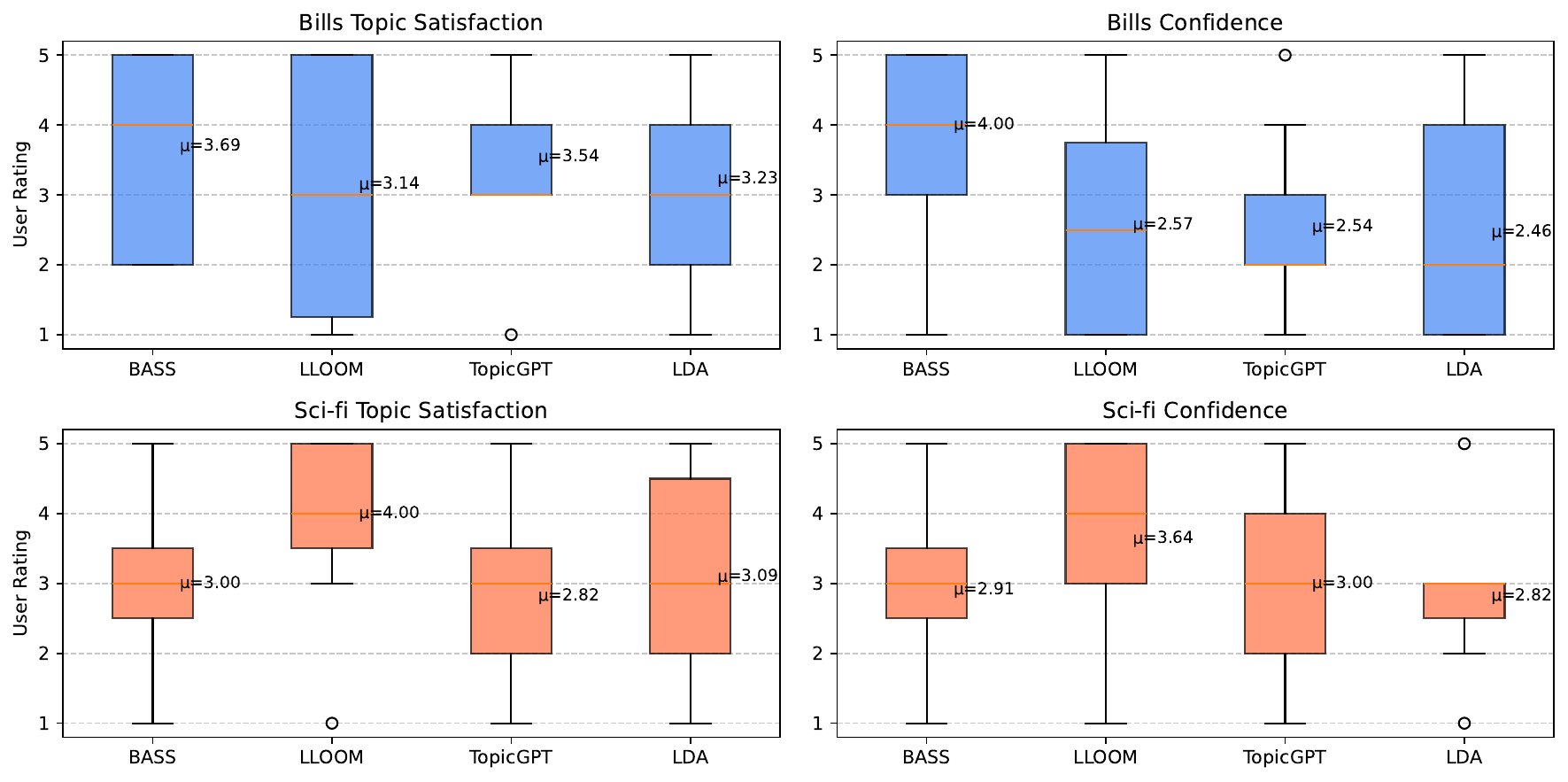}
    \caption{\bass{} has the highest user satisfaction and confidence on \texttt{Bills}, and \lloom{} has the highest ratings on \texttt{Sci-fi}. Users need to spend more mental efforts to complete the task using \abr{lda} and results in lower satisfaction rate, but the final quality of data exploration answers do not vary much from that of \mm{} models~\ref{fig:automatic_evaluation}.}
    \label{fig:survey_ratings}
\end{figure}

\section{Topic Modeling in Domain Specific Datasets}
\label{sec:domain_specific_data}
We show topic model outputs on three domain-specific datasets.
\begin{itemize}
    \item National Center for Teacher Effectiveness (\abr{ncte}): a teacher and student conversation in a math classroom to access teaching practices associated with overall high-quality math teaching~\cite{xu2024promises}. 
    \abr{ncte} has gold expert defined high-level concepts that require full understanding of education, teaching, and the dataset to derive, not just based on word frequencies-- \textit{Mathematical Language: captures how fluent teacher and students use mathematical language in a classroom}.
    \item Synthetic \texttt{Sci-Fi}: a synthetic news dataset generated by the authors about aliens science fiction. Example topics are \textit{Cultural and societal implications: Examining how humanity's institutions, values, and norms might be affected by contact with an alien intelligence}.
    \item Mathematics Aptitude Test of Heuristics (\abr{math}): a math competition question dataset involves AMC 10, AMC 12 with full step solutions and explanations~\cite{hendrycksmath2021}.
\end{itemize}

See Table~\ref{tab:topic_generation_comparison} for example output topics from those models.

\subsection{Post User Survey}
We evaluate users' experience by asking them them survey questions 1: \textit{How satisfied are you on topics that you use to answer the post test questions?} 2: \textit{How confident are you in the quality of answer you put after exploring the data with the tool?}

All the questions aim to understand the usefulness of topics for each method in helping users explore and understand essential contents in a dataset.
We plot the user reported ratings in Fig.~\ref{fig:survey_ratings}. 
%
% One interesting thing is that what users feel about the tool is not what help them get the best answers. 
% %
% For example, \lloom{} users have high satisfactions and confidence on the \texttt{Sci-fi} data, 
% \zongxiacomment{Discuss the potential problems.}
Overall, users are more satisfied and confident with their answers when an \mm{} is involved than \abr{lda}.

\subsection{Example Generated Topics on Domain Specific Data}
\label{app:exmaple_generated_topics}
Table~\ref{tab:topic_generation_comparison} shows example generated topics on domain specific data using different \mm{}s. 
\topicgpt{} specifically struggles at generating suitable and specific enough topics for domain specific datasets.

\begin{table*}[h!]
\centering
\small
\begin{tabular}{p{0.12\textwidth}p{0.26\textwidth}p{0.26\textwidth}p{0.26\textwidth}}
\toprule
\textbf{Model} & \textbf{\abr{ncte}} & \textbf{\texttt{Sci-fi}} & \textbf{\abr{math}} \\
\midrule
\abr{lda} & 
\begin{itemize}[leftmargin=*,noitemsep,topsep=0pt]
    \item \textit{Topic 1}: `apple', `row', `thirteen', `story', `division'
    \item \textit{Topic 2}: `remainder', `row', `division', `sentence', `pencil'
    \item \textit{Topic 3}: `factor', `simple', `color', `fit', `parenthesis'
\end{itemize} &
\begin{itemize}[leftmargin=*,noitemsep,topsep=0pt]
    \item \textit{Topic 1}: `silent', `quinlan', `prime', `erebus', `vaughn'
    \item \textit{Topic 2}: `humanity', `ravage', `world', `great', `planet'
    \item \textit{Topic 3}: `crew', `ship', `hope', `spaceship', `alien'
\end{itemize} &
\begin{itemize}[leftmargin=*,noitemsep,topsep=0pt]
    \item \textit{Topic 1}: `day', `team', `dotlinewidthbp', `girl', `mile'
    \item \textit{Topic 2}: `log', `cdot', `lfloor', `rfloor', `frac'
    \item \textit{Topic 3}: `bead', `textif', `endcase', `blue', `begincases'
\end{itemize} \\
\midrule
\topicgpt{} & 
\begin{itemize}[leftmargin=*,noitemsep,topsep=0pt]
    \item \texttt{[1]} \textit{Education}: The document discusses teaching methods and classroom interactions.
    \item \texttt{[2]} \textit{Mathematics Instruction}: Discusses teaching methods and student interactions in a mathematics classroom.
    \item \texttt{[2]} \textit{Classroom Management}: Discusses teacher-student interactions and classroom dynamics.
\end{itemize} &
\begin{itemize}[leftmargin=*,noitemsep,topsep=0pt]
    \item \textcolor{red}{\texttt{[1]} \textit{Science and Technology}: Involves the study and application of scientific and technological advancements.}
    \item \texttt{[2]} \textit{Non-Human Intelligence}: Mentions encounters and interactions with non-human intelligences, their behaviors, and the implications for humanity.
    \item \texttt{[2]} \textit{Interspecies Communication}: Discusses the challenges and methodologies of establishing communication with non-human intelligences.
\end{itemize} &
\begin{itemize}[leftmargin=*,noitemsep,topsep=0pt]
    \item \textcolor{red}{\texttt{[1]} \textit{Agriculture}: Mentions policies relating to agricultural practices and products.}
\end{itemize} \\
\midrule
\lloom{} & 
\begin{itemize}[leftmargin=*,noitemsep,topsep=0pt]
    \item \textit{Volume and Dimensions}: Is the focus of this text on calculating volume and understanding dimensions?
    \item  \textit{Symmetry and Shapes}: Is this text about identifying symmetry in shapes or using shapes to teach symmetry?
    \item \textit{Real-world Math}: Does this example integrate math concepts into real-world scenarios or problems?
\end{itemize} &
\begin{itemize}[leftmargin=*,noitemsep,topsep=0pt]
    \item \textit{Reality Manipulation}: Is reality manipulation or alteration a key theme in this text?
    \item \textit{Interspecies Communication}: Does the example involve efforts or challenges in communicating with a different species or entity?
    \item \textit{Existential Reevaluation}: Does this text describe a scenario that leads to an existential crisis and a reevaluation of human values or society?
\end{itemize} &
\begin{itemize}[leftmargin=*,noitemsep,topsep=0pt]
    \item \textit{Complex Numbers}: Does this example deal with complex numbers or their properties?
    \item \textit{Probability and Statistics}: Does this example involve calculating probabilities, statistical analysis, or outcomes of random events?
    \item \textit{Divisibility and Primes}: Does this example deal with factors, multiples, divisibility rules, or properties of prime numbers?
\end{itemize} \\
\midrule
\bass{} & 
\begin{itemize}[leftmargin=*,noitemsep,topsep=0pt]
    \item \textit{Mathematics education}: The teacher employs a method of engaging students through continuous questioning and prompting them to explain their reasoning. This interactive approach helps students articulate their thought processes and understand the concepts being discussed.
    \item \textit{Interactive questioning}: The document showcases a teaching strategy where the teacher uses a game-based approach to teach addition and number sense. 
    \item \textit{Multiplication and division}: The teacher prompts students to explain their strategies, whether they have memorized facts or used other methods, and guides them through the process of writing multiplication and division sentences
\end{itemize} &
\begin{itemize}[leftmargin=*,noitemsep,topsep=0pt]
    \item \textit{Relations between extra terrestrial and humanity}: The document revolves around the discovery of an alien signal, the subsequent decoding of messages from an ancient intelligence, and the ethical and moral implications of engaging with this non-human entity...
    \item \textit{universe challenges humanitys understanding of existence}: The document focus is on the interaction and communication between humans and an alien species, as well as the societal structures and knowledge exchange.
    \item \textit{interdimensional exploration and politics}: The document describes a scenario where humanity, under the Roman Empire, explores and interacts with a non-human intelligence across multiple realities...
\end{itemize} &
\begin{itemize}[leftmargin=*,noitemsep,topsep=0pt]
    \item \textit{Probability and prime numbers}: involves calculating the probability of a specific event involving prime numbers, which falls under the study of probability and prime numbers.
    \item \textit{Geometry}: involves geometric properties and relationships within a triangle.
    \item \textit{Number theory}: The topic involves the concepts of greatest common divisor (gcd) and least common multiple (lcm), which are fundamental topics in number theory.
    \item \textit{Polynomial equations analysis}: The topic involves solving an algebraic equation with specific conditions related to its roots.
\end{itemize} \\
\bottomrule
\end{tabular}
\caption{Generated topics for different datasets across various models. For \topicgpt{}, \texttt{[1]} means the first level topics, and \texttt{[2]} means second level topics. \topicgpt{} appears to hallucinate few first-level topics on domain-specific data.}
\label{tab:topic_generation_comparison}
\end{table*}

% \section{Domain-Specific Data}

\section{Generation of the Synthetic dataset}\label{appendix:synthetic}
Algorithm~\ref{alg:scifi_alg} contains the pseudo-code used to generate our synthetic datasets, while Prompt~\ref{prompt:scifi_system} provides the system prompt, and Prompt~\ref{prompt:scifi_user} gives an example of a user prompt.

\onecolumn

\begin{algorithm}[!h]
\caption{Generate Synthetic Dataset (\texttt{Sci-Fi)}}
\label{alg:scifi_alg}
\begin{algorithmic}[1]

\State \textbf{Input:} Sets of styles $S$, themes $T$, settings $G$, moods $M$, and question-answer pairs $Q$
\State \textbf{Output:} Response text and input parameters stored in an output file

% Generate each user prompt combinations 
\State Initialize $U$ (user prompt combinations) as an empty list
\State Compute the set $P = \{ (k_1, k_2) \mid k_1, k_2 \in K, k_1 \neq k_2 \}$
\For{each $(s, (k_1, k_2), g) \in S \times P \times G$}
    \State Select a random mood $m \in M$
    \State Select a random $(q, a) \in Q$
    \State Add $(s, k_1, k_2, g, m, q, a)$ to $U$
\EndFor

% Randomly shuffle the user prompt combinations
\State Randomly shuffle $U$

% Initialize common words dictionary...
\State Initialize $D$ (common words dictionary) as an empty dictionary
\State Load sample text $T_s$
\State Tokenize $T_s$ into words $W$
\For{each $w \in W$}
    \State Strip punctuation and possessives from $w$
    \If{$|w| > 4$ AND $w$ starts with a capital letter AND $w \notin$ stop words}
        \State Add $w$ to $D$
    \EndIf
    \If{$w$ contains four consecutive digits}
        \State Add $w$ to $D$
    \EndIf
\EndFor
\State Add predefined opening words $\{\text{``In the''}, \text{``On the''}\}$ to $D$

% For each user prompt parameter combination:
\For{each $(s, k_1, k_2, g, m, q, a) \in U$}
    \State Define $W_{\text{avoid}}$ as the 250 most common words in $D$
    \State Generate $P_u$ (user prompt) using a predefined template with $(s, m, k_1, k_2, g, q, a, W_{\text{avoid}})$
    \State Send system and user prompts to LLM
    \State Receive response text $R$ from LLM

    % same process as when creating the dict in the prev step
    \State Tokenize $R$ into words $W_r$
    \For{each $w \in W_r$}
        \State Strip punctuation and possessives from $w$
        \If{$|w| > 4$ AND $w$ starts with a capital letter AND $w \notin$ stop words}
            \State Add $w$ to $D$
        \EndIf
        \If{$w$ contains four consecutive digits}
            \State Add $w$ to $D$
        \EndIf
    \EndFor
    
    \State Write $(R, s, m, k_1, k_2, g, q, a)$ to output file
\EndFor

\end{algorithmic}
\end{algorithm}

\newpage

\begin{prompt}[title={\thetcbcounter: Prompt for generating topic suggestions for \bass{}}, label=fig:bass_prompt]
\begin{lstlisting}
You will receive a document about the congressional Bills and a set of top-level topics from a topic hierarchy. Your task is to identify an policy topic within the document that can act as top-level topics in the hierarchy. If any relevant topics are missing from the provided set, please add them. Otherwise, output the existing top-level topics as identified in the document.

Follow the following format:

DOCUMENT: [DOCUMENT]

HIGH LEVEL CONCEPTS: [HIGH\_LEVEL\_CONCEPTS]

YOU SHOULD STRICTLY FOLLOW THE FOLLOWING FORMAT AND OUTPUT THE FOLLOWING INFORMATION

RATIONALE: Rationale for choosing the high-level concept

PRED CONCEPT: High-level concept for the document

----------

Previous USER LABELED EXAMPLES (AT MOST THREE) IF AVAILABLE

----------

DOCUMENT: {}  HIGH LEVEL CONCEPTS: {}  

As a reminder, you should output the following information following the given output format. Your generated concept should not EXCEED FIVE words. Your generated concept should be the teacher's teaching strategy, not a general theme such as 'Education'

RATIONALE: Your rationale for making such a label

PRED CONCEPT: Your generated concept
\end{lstlisting}
\end{prompt}

\begin{prompt}[title={\thetcbcounter: System Prompt for Sci-Fi Generation}, label=prompt:scifi_system]
\begin{lstlisting}
You are a clever research assistant generating synthetic data for a human subject study.

Using the style, mood, themes, setting, question/answer pair and the list of words to avoid provided to you in the user prompt, create a Wikipedia-style full plot summary of a science fiction story about first contact with a non-human intelligence including spoilers and the final plot resolution. 

Fastidiously adhere to following rules:
* Use the question/answer pair to provide the reader with descriptive information about the non-human-intelligence, but do not reveal the question in the generated text.
* Avoid cliche openings like: 'In the', 'Within the', 'On the', 'As the'
* Don't the use the words in the user provided list of words to avoid.
* Be creative in your choices of proper names for people, places, and entities.
* Don't choose a word to avoid as a proper name.
* Be creative in your choices of dates.
* Don't choose a year from the words to avoid.
* Do not start the summary with a title. 
* Do not directly reveal the themes to the reader in your generated text. 
* Be sure to emphasize the theme, but do not ask questions in your plot summary. 
* Do not preface your response with statments like: "Here is a Wikipedia-style science fiction plot summary:\n\n\" or make other statements suggesting that the output is generated.
\end{lstlisting}
\end{prompt}
\clearpage
\begin{prompt}[title={\thetcbcounter: User Prompt example for Sci-Fi Generation}, label=prompt:scifi_user]
\begin{lstlisting}
Style: Hard Science Fiction: This style focuses on scientific accuracy and technical details, often featuring engineers, scientists, and inventors as main characters. Examples: Isaac Asimov, Arthur C. Clarke, and Kim Stanley Robinson.
Mood: hopeful
Theme 1: The Other: Exploring the nature of the alien intelligence, its motivations, and its place in the universe.
Theme 2: Humanity's place in the universe: Questioning humanity's significance, morality, and purpose in the face of a non-human intelligence.
Setting: Space stations or colonies: Isolated and vulnerable, these settings can heighten the sense of tension and uncertainty.
Question: What challenges did the humans face when trying to communicate with the non-human intelligence?
Answer: Understanding the non-human intelligence's  motivations and intentions that are fundamentally different from human principles, making it difficult to comprehend its actions and goals. 
\end{lstlisting}
\end{prompt}

\twocolumn

 \label{appendix}

\end{document}